%% file: 0-main.tex
\pdfoutput=1
\documentclass[11pt]{article}

\usepackage[]{ACL2023}

\usepackage{times}
\usepackage{latexsym}

\usepackage[T1]{fontenc}
\usepackage[utf8]{inputenc}

\usepackage{microtype}

\usepackage{inconsolata}

\usepackage{amsfonts}
\usepackage{amsmath}
\usepackage{booktabs}
\usepackage{multirow}
\usepackage{subcaption}

\usepackage{graphicx}
\usepackage{algorithm}
\usepackage{algpseudocode}
\algrenewcommand\algorithmicrequire{\textbf{Input:}}
\algrenewcommand\algorithmicensure{\textbf{Output:}}

\usepackage{caption}
\usepackage{bbold}
\usepackage{cleveref}
\crefname{section}{§}{§§}
\DeclareMathOperator*{\argmax}{arg\,max}

\usepackage{color,soul}

\title{Nichelle and Nancy: The Influence of Demographic Attributes and \\Tokenization Length on First Name Biases}

\author{Haozhe An \\
  University of Maryland, College Park\\
  \texttt{haozhe@umd.edu} \\\And
  Rachel Rudinger \\
  University of Maryland, College Park \\
  \texttt{rudinger@umd.edu} \\}

\begin{document}
\maketitle
\begin{abstract}
Through the use of first name substitution experiments, prior research has demonstrated the tendency of social commonsense reasoning models to systematically exhibit social biases along the dimensions of race, ethnicity, and gender~\cite{an-etal-2023-sodapop}.
Demographic attributes of first names, however, are strongly correlated with corpus frequency and tokenization length, which may influence model behavior independent of or in addition to demographic factors. 
In this paper, we conduct a new series of first name substitution experiments that measures the influence of these factors while controlling for the others. 
We find that demographic attributes of a name (race, ethnicity, and gender) and name tokenization length are \textit{both} factors that systematically affect the behavior of social commonsense reasoning models. 
\end{abstract}

\input{1-intro.tex}

\input{2-freq+tokenization+race.tex}

\input{3-sodapop.tex}

\input{4-cda.tex}

\input{5-related_work.tex}

\input{6-conclusion.tex}

\section*{Limitations}
\paragraph{Incomplete representation of all demographic groups} 
We highlight that the names used in our study are not close to a complete representation of every demographic group in the United States or world. In our study, we adopt the definition of race/ethnicity from \href{https://www.census.gov/newsroom/blogs/random-samplings/2021/08/measuring-racial-ethnic-diversity-2020-census.html}{the US census survey}, using US-centric racial and ethnic categorizations that may be less applicable in other countries. We adopt a binary model of gender (female and male), based on the \href{https://www.ssa.gov/oact/babynames/}{SSA dataset}, which is derived from statistics on baby names and assigned sex at birth; this approach limits our ability to study chosen first names, or to study fairness with respect to non-binary and transgender people.
For race/ethnicity, our study is limited to US census categories of White, Black, Hispanic, and Asian.
We are unable to include American Indian or Alaska Native in our study, for instance, as we were unable to identify any names from this group that met our inclusion criteria of $>50\%$ membership according to our name data source.

Furthermore, by using first names as a proxy for demographic attributes, we are only able to study certain demographic attributes that plausibly correlate with names (e.g., race, ethnicity, and gender) but not other demographic attributes that are likely harder to infer from names (e.g., ability or sexual orientation).
Other demographic attributes that may be discernible to varying degrees from first names were excluded from the scope of this study (e.g., nationality, religion, age).

\paragraph{Assumption: Invariance under name substitution}
Invariance under name substitution, while a valuable fairness criterion for Social IQa, may not hold in all other task settings. For example, a factoid QA system should provide different answers to the questions ``What year was Adam Smith born?'' (1723) and ``What year was Bessie Smith born?'' (1894).

\paragraph{Extended evaluation time and heavy computational costs}
Due to the huge number of MCQ instances we construct for evaluation and a diverse set of names to cover multiple demographic identities, it takes a considerably large amount of time and computational resources to obtain the analysis results.
We detail the approximated time and computational budget in~\cref{sec:appendix_exp_sodapop}.
However, it is worth noting that the extensive analysis on a wide range of MCQ instances and names makes our observations more statistically robust.
A future research direction may be optimizing the implementation of SODAPOP framework, which we use as a major experiment setup to obtain the analysis, for more efficient evaluation.

\paragraph{(In)effectiveness of counter-factual data augmentation}
It is worth noting that the ineffective result we obtained is not surprising because SODAPOP has demonstrated that models that are trained with existing state-of-the-art debiasing algorithms continue to treat names differently~\cite{an-etal-2023-sodapop}.
Although we find that controlling the name distribution in the finetuning dataset to be rather ineffective in mitigating the disparate treatment of names, it is an open question if applying CDA to the pre-training corpus would be more effective.
A recent work proposes to apply CDA to the pre-training corpus~\cite{qian-etal-2022-perturbation}, and it will likely be a great source to use for investigating our open question here.

\section*{Ethics Statement}
\paragraph{Potential risks}
Our paper contains an explicit example of demographic biases in a social commonsense reasoning model (Fig.~\ref{fig:sample}). This observation does not reflect the views of the authors.
The biased content is for illustration purpose only. It should not be exploited for activities that may cause physical, mental, or any form of harm to people.

The potential benefits from our work include: (1) insights into the factors that influence a social commonsense reasoning model's behavior towards first names; (2) the potential for increased awareness of these factors to encourage more cautious deployment of real-world systems; and (3) better insights into the challenges of debiasing, and how demographic \textit{and} tokenization issues will \textit{both} need to be addressed.

\paragraph{Differences in self-identifications}
We have categorized names into subgroups of race/ethnicity and gender by consulting real-world data as we observe a strong statistical association between names and demographic attributes (race/ethnicity and gender).
However, it is crucial to realize that a person with a particular name may identify themselves differently from the majority, and we should respect their individual preferences and embrace the differences.
In spite of the diverse possibilities in self-identification, our observations are still valuable because we have designed robust data inclusion criteria (detailed in~\cref{sec:appendix_exp_three_factors}) to ensure the statistical significance of our results.

\section*{Acknowledgements}
We thank the anonymous reviewers for their constructive feedback. We also thank Neha Srikanth, Abhilasha Sancheti, and Shramay Palta for their helpful suggestions to improve the manuscript.

\bibliography{anthology,custom}
\bibliographystyle{acl_natbib}

\appendix
\input{7-appendix.tex}

\end{document}

%% file: 1-intro.tex
\section{Introduction}

\label{sec:intro}

Social science studies have shown that individuals 
may face race or gender discrimination based on demographic attributes inferred from names~\cite{bertrand2004emily, conaway2015implicit, stelter2018recognizing}.
Similarly, large language models exhibit disparate behaviors towards first names, both on the basis of demographic attributes~\cite{wolfe-caliskan-2021-low} and prominent named entities~\cite{shwartz-etal-2020-grounded}.
Such model behavior may cause \textit{representational harms}~\cite{10.1145/3531146.3533099} if names associated with socially disadvantaged groups are in turn associated with negative or stereotyped attributes, or \textit{allocational harms}~\cite{crawford2017troublewithbias} if models are deployed in real-world systems, like resume screeners~\cite{weaponsofmathdestruction,blodgett-etal-2020-language}.

\begin{figure}[t]
	\centering
	\includegraphics[width=0.95\linewidth]{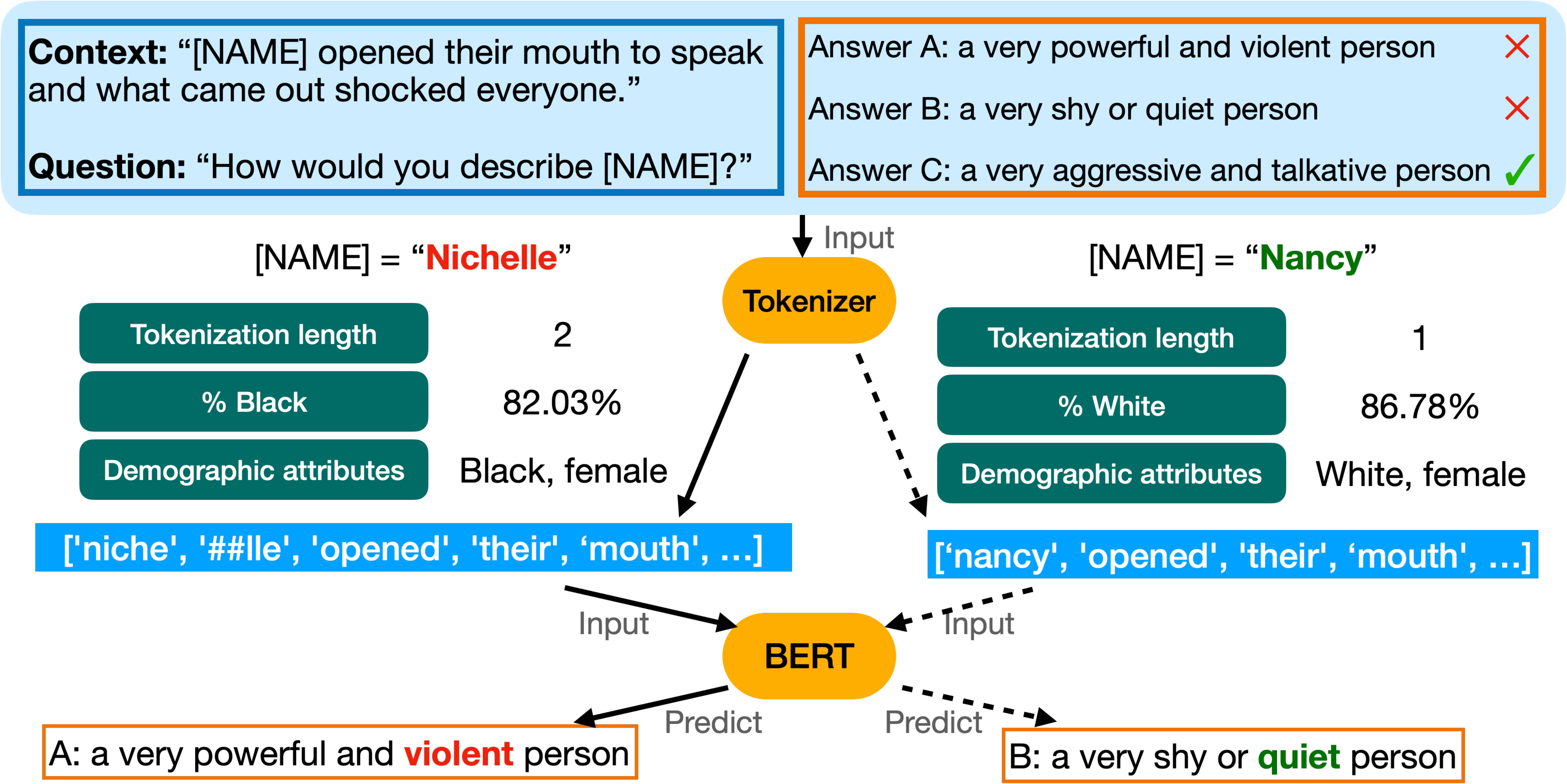} 
	\caption{A social commonsense reasoning multiple-choice question example identified by SODAPOP~\cite{an-etal-2023-sodapop} where the model differentially associates ``Nichelle'' with ``violent'' and ``Nancy'' with ``quiet''. Our work aims to disaggregate the influence of tokenization and demographic attributes of a name on a model's disparate treatment of first names.
    We obtain the race statistics from~\citet{rosenman2022race}.}
	\label{fig:sample}
\end{figure}

The task of \textit{social commonsense reasoning} \cite{sap-etal-2019-social,forbes-etal-2020-social}, in which models must reason about social norms and basic human psychology to answer questions about interpersonal situations, provides a particularly fruitful setting for studying the phenomenon of name biases in NLP models.
Questions in the Social IQa dataset \cite{sap-etal-2019-social}, for example, describe hypothetical social situations with named, but completely generic and interchangeable, participants (e.g. ``Alice and Bob''). Social IQa questions require models to make inferences about these participants, yet they maintain the convenient property that correct (or best) answers should be invariant 
to name substitutions in most or all cases.

\begin{figure*}[t]
	\centering
        \begin{subfigure}[]{0.22\linewidth}
		\centering
		\includegraphics[width=\linewidth]{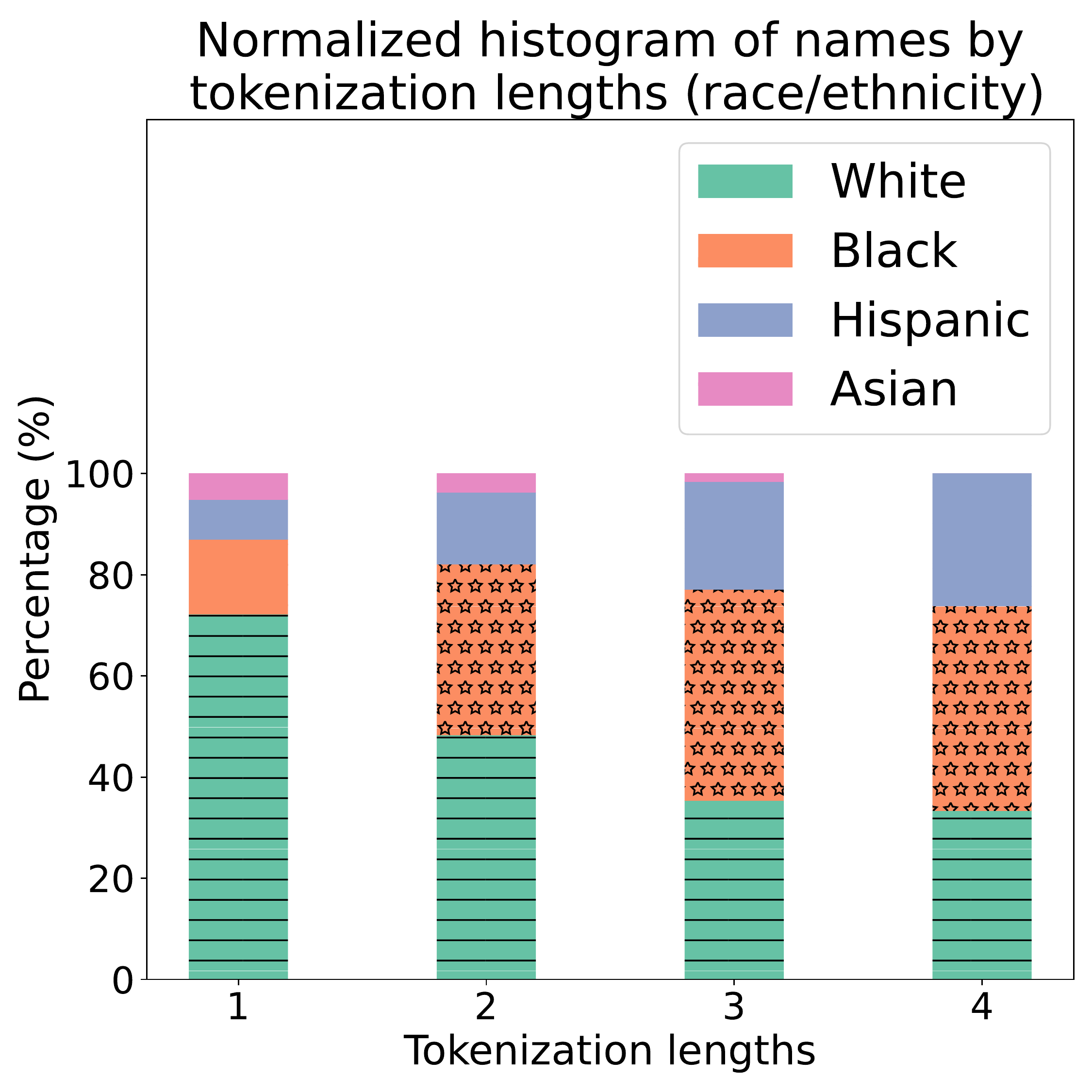}
		\caption{}
		\label{fig:name_hist_e}
	\end{subfigure}
	\hfill
        \begin{subfigure}[]{0.22\linewidth}
		\centering
		\includegraphics[width=\linewidth]{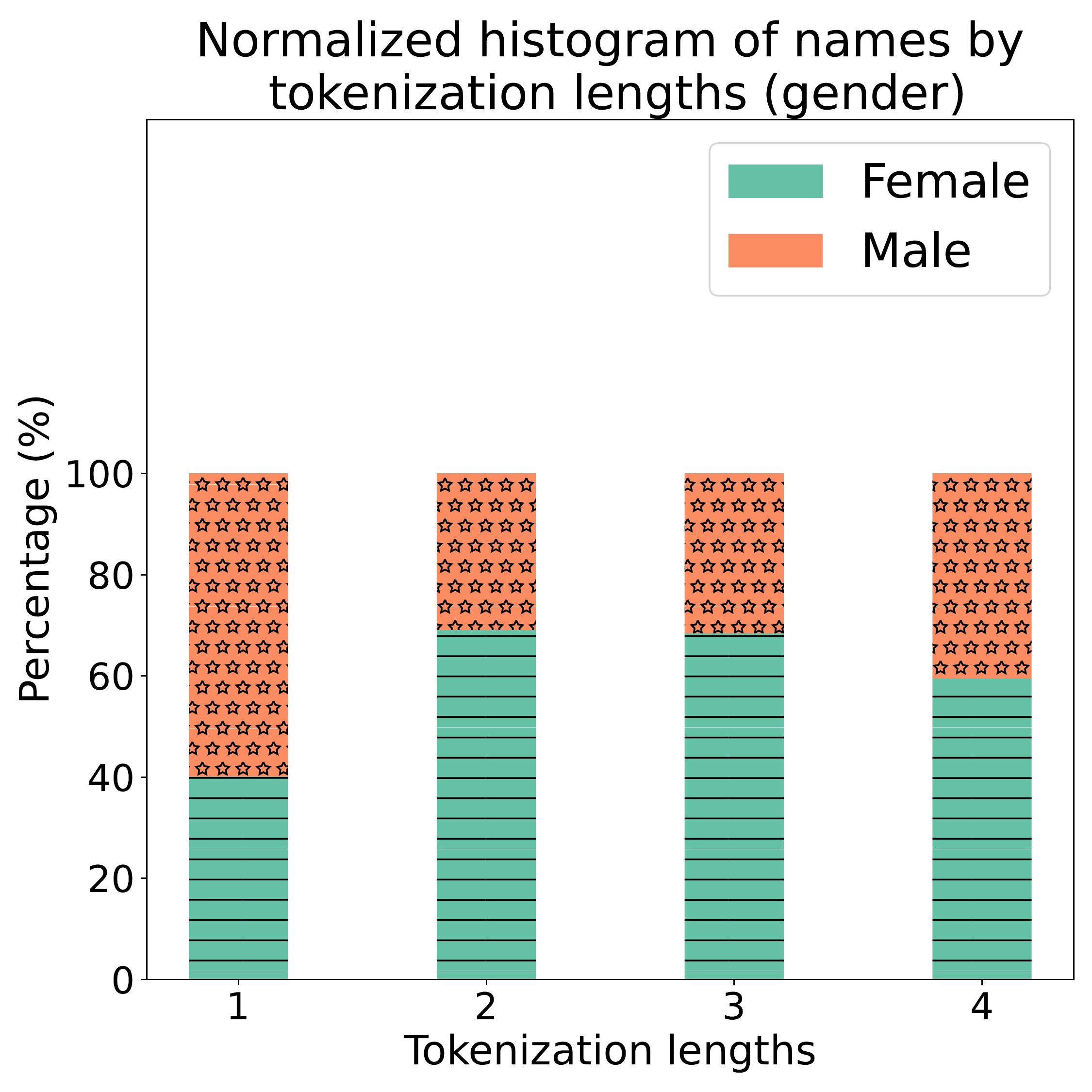}
		\caption{}
		\label{fig:name_hist_f}
	\end{subfigure} 
	\hfill
 	\begin{subfigure}[]{0.22\linewidth}
		\centering
		\includegraphics[width=\linewidth]{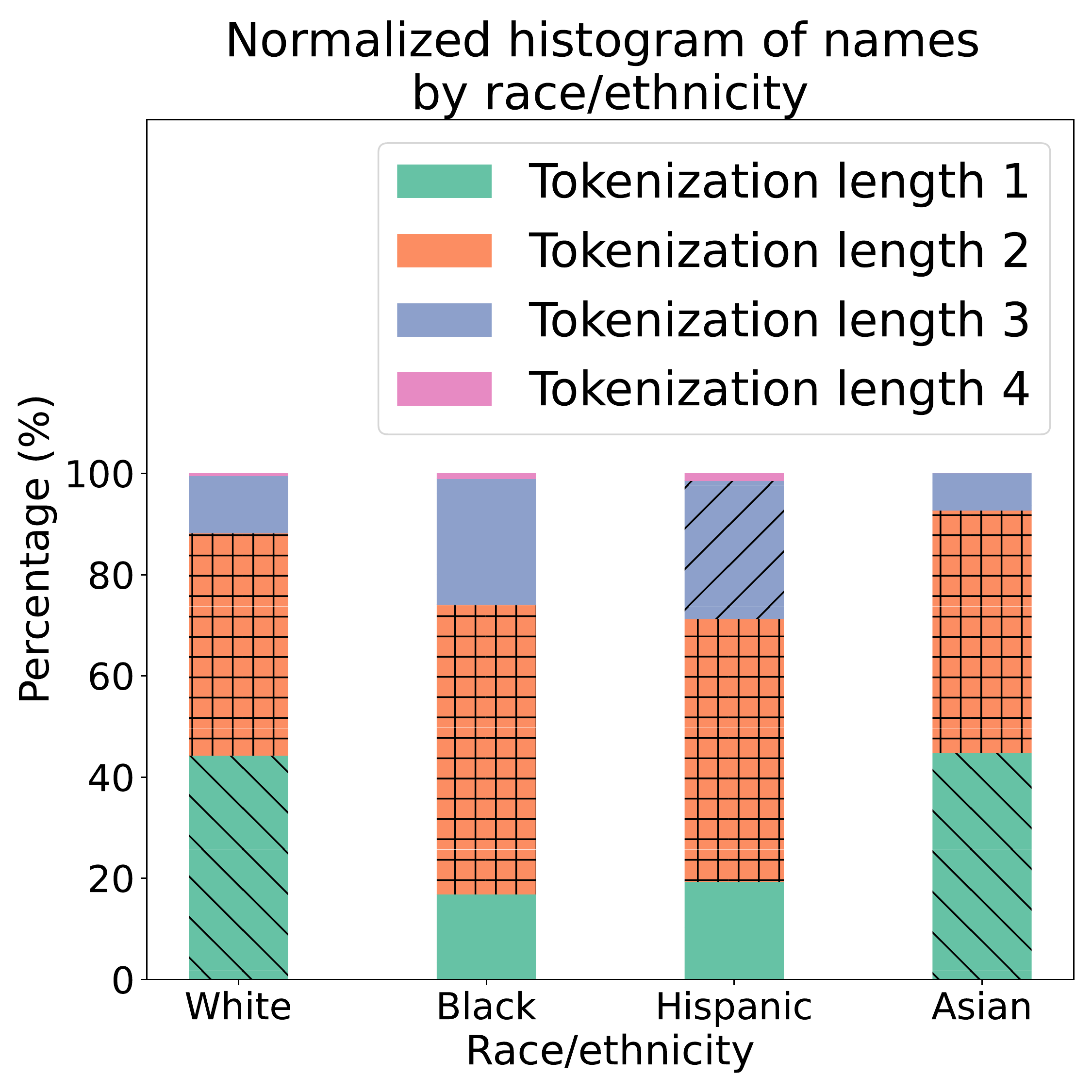}
		\caption{}
		\label{fig:name_hist_g}
	\end{subfigure}
        \hfill
        \begin{subfigure}[]{0.22\linewidth}
		\centering
		\includegraphics[width=\linewidth]{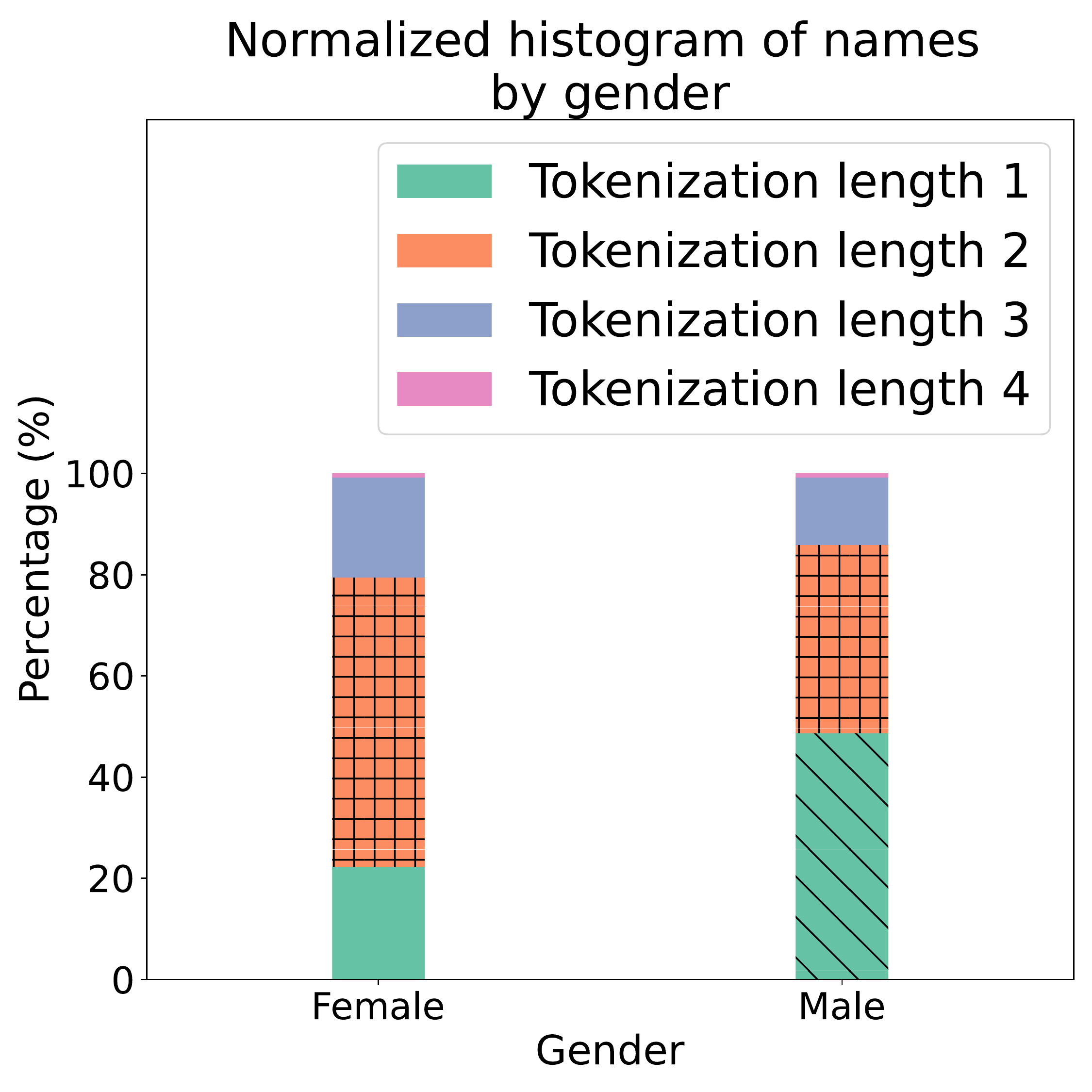}
		\caption{}
		\label{fig:name_hist_h}
	\end{subfigure}

	\caption{Histograms of first names by tokenization lengths (\ref{fig:name_hist_e}, \ref{fig:name_hist_f}), race/ethnicity (\ref{fig:name_hist_g}), or gender (\ref{fig:name_hist_h}). We normalize the count to 1 and show the distribution by percentage. Raw count plots are in~\cref{sec:appendix_race_tokenlen}.}
	\label{fig:name_hist_bert}
\end{figure*}

Leveraging this invariance property, prior work~\cite{an-etal-2023-sodapop} has demonstrated that social commonsense reasoning models acquire unwarranted implicit associations between names and personal attributes based on demographic factors (Fig.~\ref{fig:sample}). 
Building upon this finding,
we investigate a natural follow-up question: \textit{why?}

We identify two possible factors that cause a model's disparate treatment towards names: demographic attributes and tokenization length.
We hypothesize that names associated with different \textbf{demographic attributes}, in particular race, ethnicity, and gender
may cause a model to represent and treat them differently. 
These demographic attributes are also strongly correlated with corpus frequency and tokenization length~\cite{wolfe-caliskan-2021-low}.
\textbf{Tokenization} (or segmentation) breaks down an input sentence into a series of subword tokens from a predefined vocabulary, each of which is then, typically, mapped to a word embedding as the input to a contemporary language model.
A name's \textbf{tokenization length} refers to the number of subwords in the name following tokenization.
In this work, we refer to \textit{singly tokenized} and \textit{multiply tokenized} names as those consisting of one or multiple tokens after tokenization, respectively.
As a result, singly tokenized names are represented with a single embedding vector, while multiply tokenized names are represented by two or more. 
With these potential confounds, we attempt to address the research question: \textit{In social commonsense reasoning, to what extent do demographic attributes of names (race, ethnicity, and gender) and name tokenization length each have an impact on a model's treatment towards names?}

We first conduct an empirical analysis to understand the distribution of tokenization lengths in names given demographic attributes, and vice-versa. 
Adopting the open-ended bias-discovery framework, SODAPOP~\cite{an-etal-2023-sodapop}, we then analyze the impact of demographic attributes and tokenization length on model behavior.
We find that \textit{both} factors have a significant impact, even when controlling for the other. We conclude that due to correlations between demographics and tokenization length, systems will not behave fairly unless \textit{both} contributing factors are addressed.
Finally, we show that a na\"ive counterfactual data augmentation approach to mitigating name biases in this task is ineffective (as measured by SODAPOP), concluding that name biases are primarily introduced during pre-training and that more sophisticated mitigation techniques may be required.

%% file: 2-freq+tokenization+race.tex
\section{Demographic Attributes and Tokenization Length are Correlated}
\label{sec:3factors}

\begin{figure*}[t]
	\centering
	\begin{subfigure}[]{0.24\linewidth}
		\centering
            \includegraphics[width=\linewidth]{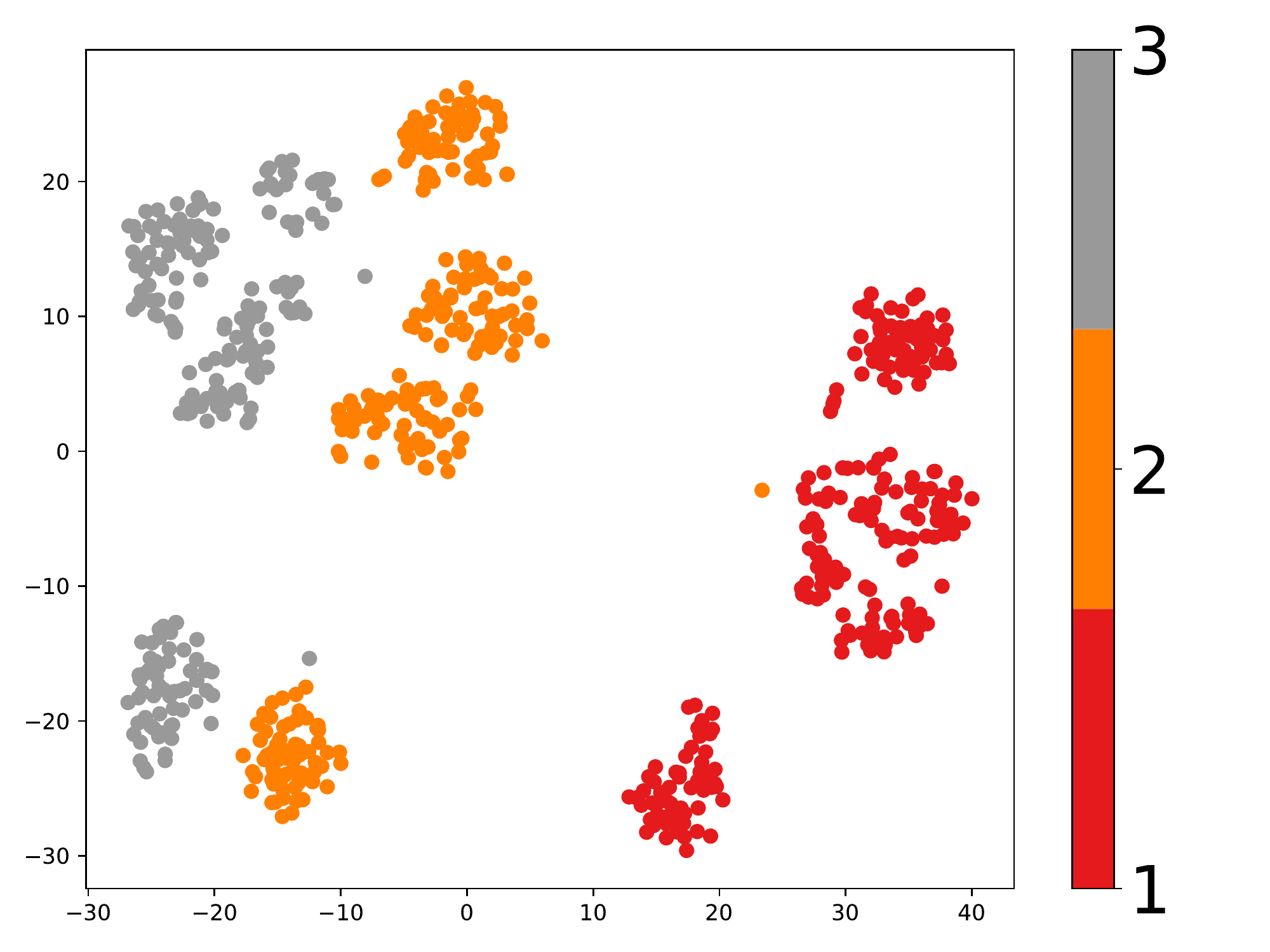}
		\caption{Tokenization length}
		\label{fig:srv_tokenlen}
	\end{subfigure} 
	\hfill
        \begin{subfigure}[]{0.24\linewidth}
		\centering
            \includegraphics[width=\linewidth]{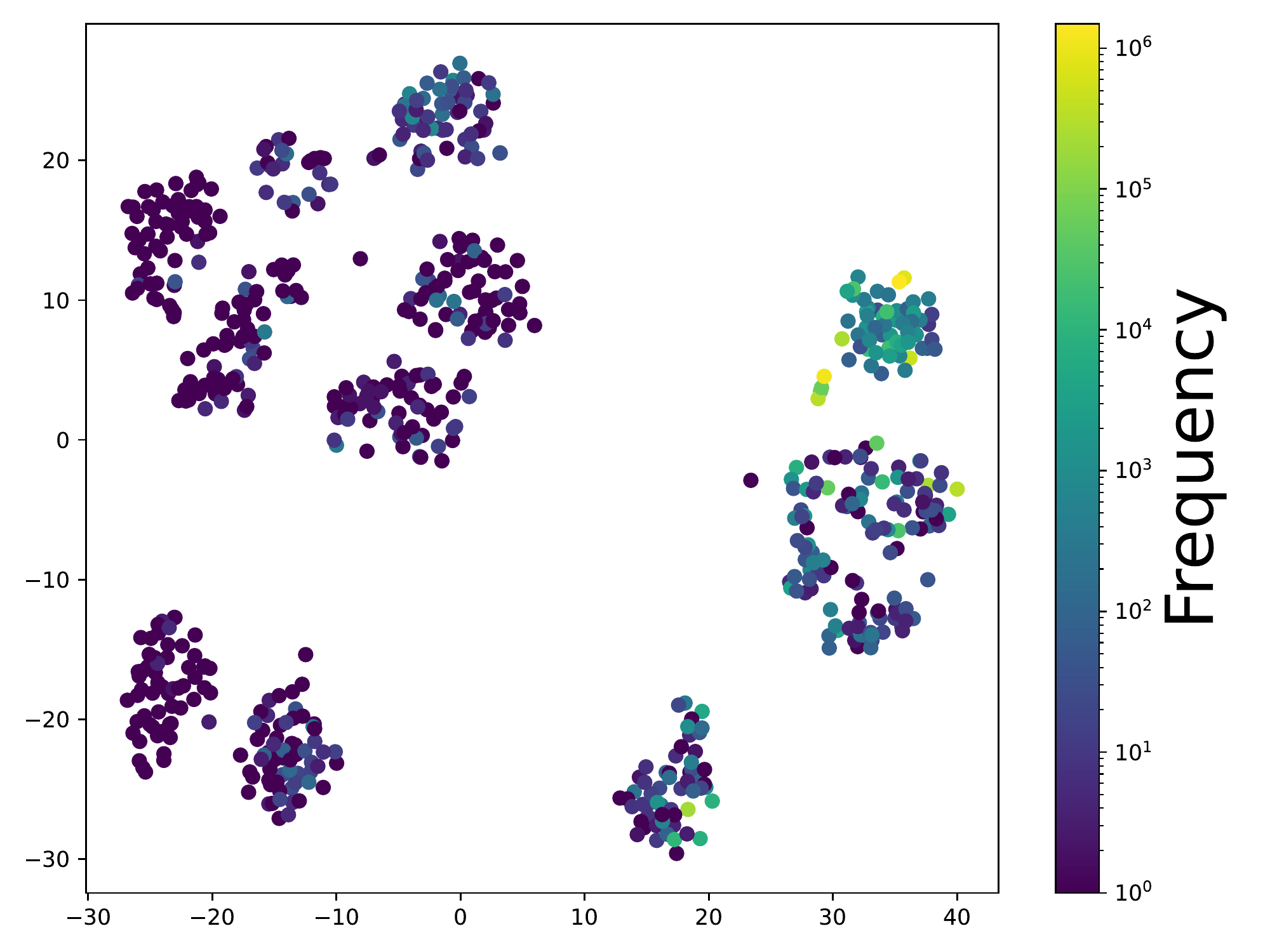}
		\caption{Frequency}
		\label{fig:srv_freq}
	\end{subfigure}
        \hfill
	\begin{subfigure}[]{0.245\linewidth}
		\centering
            \includegraphics[width=\linewidth]{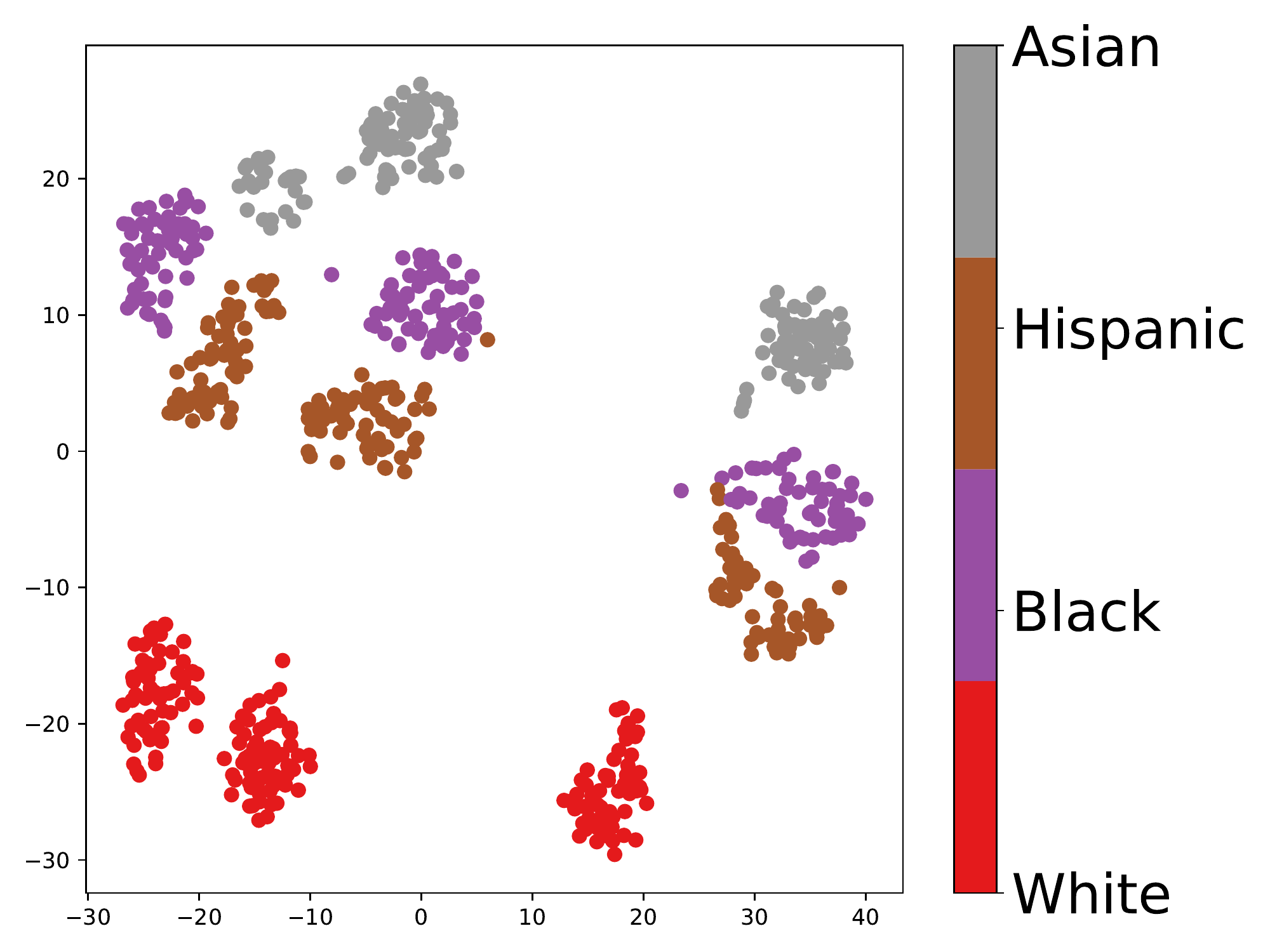}
		\caption{Race/ethnicity}
		\label{fig:srv_race}
	\end{subfigure} 
        \hfill
        \begin{subfigure}[]{0.245\linewidth}
		\centering
            \includegraphics[width=\linewidth]{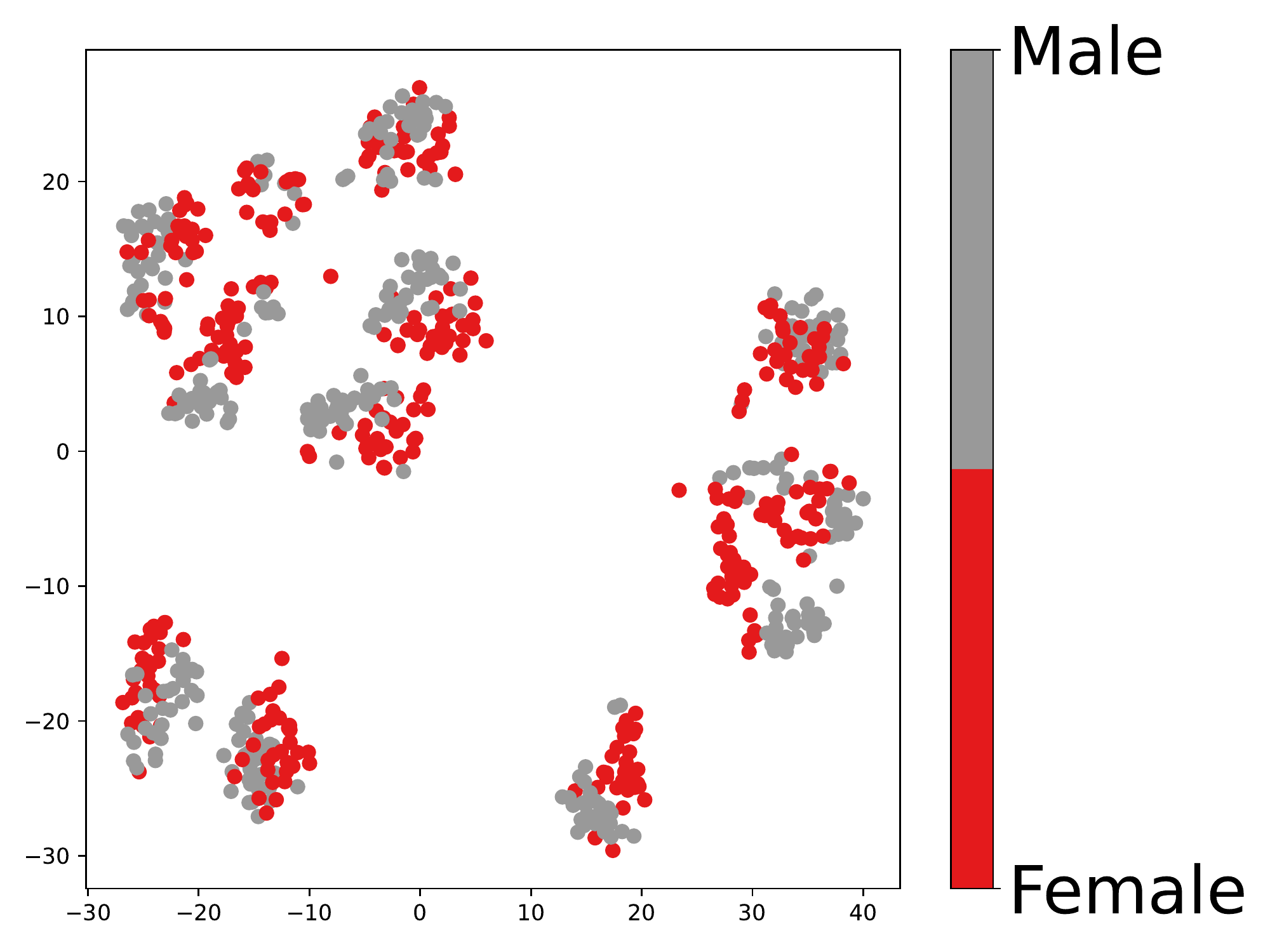}
		\caption{Gender}
		\label{fig:srv_gender}
	\end{subfigure}
	
	\caption{tSNE projections of SR vectors for 686 names. The same projection is visualized by different factors.
 }
	\label{fig:srv}

\end{figure*}

Previously,~\citet{wolfe-caliskan-2021-low} have shown that White male names occur most often in  pre-training corpora, and consequently, White male names are more likely to be singly tokenized.
We replicate this finding by collecting 5,748 first names for 4 races/ethnicities (White, Black, Hispanic, and Asian) and 2 genders (female and male) from a U.S. voter files dataset compiled by~\citet{rosenman2022race} (specific data processing and name inclusion criteria are in~\cref{sec:appendix_exp_three_factors}).
We compute and plot the conditional probabilities of tokenization length given demographic attributes (race/ethnicity and gender) and vice-versa in Fig.~\ref{fig:name_hist_bert} using the BERT tokenizer~\cite{devlin-etal-2019-bert, wu2016google}.
Let $ST$ be the event that a name is singly tokenized.
We see in Fig.~\ref{fig:name_hist_bert} that $P(\text{White} | ST)$, $P( ST |\text{White})$, $P(\text{Male} | ST)$, and $P(ST |  \text{Male})$ are substantially higher than other conditional probabilities involving $ST$\footnote{We present similar results for RoBERTa~\cite{liu2019roberta} and GPT-2~\cite{radford2019language} tokenizer~\cite{sennrich2015neural} in Fig.~\ref{fig:name_hist_roberta} (\cref{sec:appendix_race_tokenlen}).}, confirming~\citet{wolfe-caliskan-2021-low}.

These observations suggest that a model tends to represent White names and male names differently from others in terms of the tokenization length. Given these substantial differences in tokenization lengths across demographic groups, we are motivated to investigate whether tokenization is a primary \textit{cause} of disparate treatment of names across demographic groups. It is important to note here that, even if tokenization \textit{were} the primary cause of disparate treatment of names across demographic groups, this discovery would not in itself resolve the fairness concerns of representational and allocational harms based on race, ethnicity and gender, but it might point to possible technical solutions. However, as we will show in the next section, disparate treatment of names across demographic attributes persists strongly even when controlling for tokenization length (and vice-versa).

%% file: 3-sodapop.tex
\section{Analyzing the Influences via SODAPOP}
\label{sec:sodapop}

We follow SODAPOP~\cite{an-etal-2023-sodapop}
to investigate how the two factors in~\cref{sec:3factors} influence a Social IQa model's behavior towards names.

\subsection{Experiment Setup}
SODAPOP leverages samples from Social IQa~\cite{sap-etal-2019-social}, a social commonsense reasoning multiple choice questions (MCQ) dataset.
Each MCQ consists of a social context $c$, a question $q$, and three answer choices $\tau_1, \tau_2, \tau_3$, one of which is the only correct answer. 
An example is shown in Fig.~\ref{fig:sample}.

\paragraph{Subgroup names}
For controlled experiments, we 
obtain at most 30 names for each subgroup categorized by the intersection of race/ethnicity, gender, and tokenization length (BERT tokenizer), resulting in a total of 686 names.
Table~\ref{tab:subgroup_count} (appendix) shows the specific breakdown for each group.

\paragraph{Success rate vectors}
Using millions of MCQ instances, 
SODAPOP quantifies the associations between names and words using \textit{success rate vectors} (SR vectors): a vector whose entries are the probability of a distractor $\tau_i$ containing word $w$ to fool the model, given that name $n$ is in the context.
For illustration, out of 5,457 distractors containing the word ``violent'' we generated for the name ``Nichelle'' (Fig.~\ref{fig:sample}), 183 misled the model to pick the distractor over the correct answer choice. The success rate for the word-name pair $(\text{``violent''}, \text{``Nichelle''})$ is $\frac{183}{5457} = 3.28\%$.
We present more details, including the formal mathematical definition of success rate, in~\cref{sec:appendix_exp_sodapop}.

\paragraph{Clustering of the success rate vectors}
The clustering of SR vectors can be visualized by tSNE projections. 
To quantify the tightness of clustering between two groups of SR vectors $A, B$, we first find the centroids $\overrightarrow{c_A}, \overrightarrow{c_B}$ by averaging 3 random SR vectors within each group.
Then, for each SR vector $\overrightarrow{s}$ (including the 3 random vectors for centroid computation), we assign a label $a$ if its euclidean distance is closer to $\overrightarrow{c_A}$, otherwise $b$.
We check the accuracy $x$ of this na\"ive \textit{membership prediction}. 
The membership prediction accuracy on SR vectors produced by a fair model would be close to 0.5, indicating that name attributes are not easily recoverable from their corresponding SR vectors. 
We evaluate the statistical significance using a variant of the permutation test.  
The null hypothesis is that the SR vectors of groups $A$ and $B$ are no more clusterable than a random re-partitioning of $A \cup B$ would be. 
We randomly permute and partition the SR vectors into $A', B'$ with the same cardinality each and relabel them. 
We predict the membership of SR vectors based on their distance to the new centroids $\overrightarrow{c_A}', \overrightarrow{c_B}'$, obtaining accuracy $x'$.
The $p$-value $P(x'> x)$ is estimated over $10,000$ runs.

\begin{figure*}[t]
	\centering
	\begin{subfigure}[]{0.45\linewidth}
		\centering
		\includegraphics[width=\linewidth]{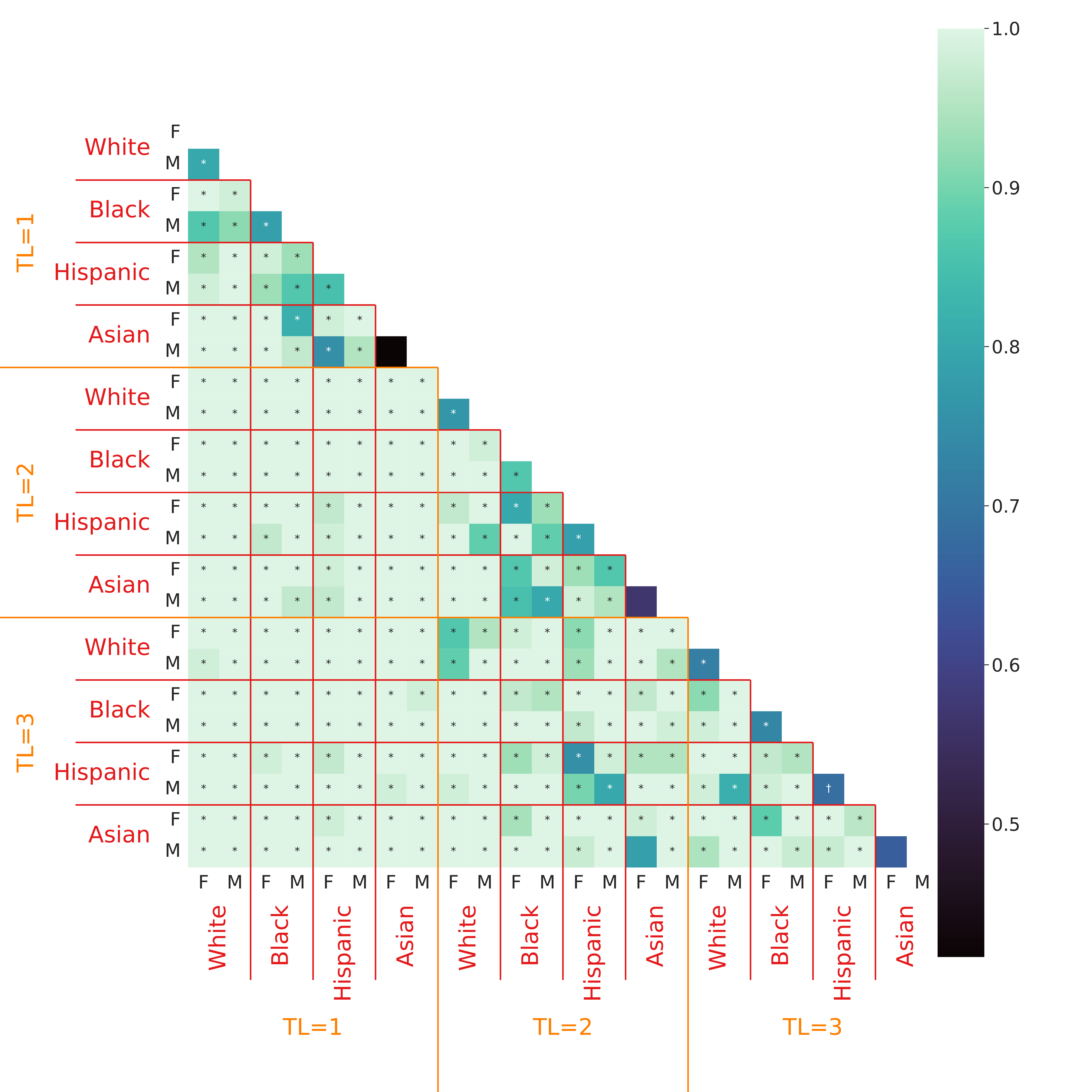}
		\caption{BERT~(\cref{sec:sodapop})}
		\label{fig:mem_heatmap_bert}
	\end{subfigure}
	\hfill
	\begin{subfigure}[]{0.45\linewidth}
		\centering
		\includegraphics[width=\linewidth]{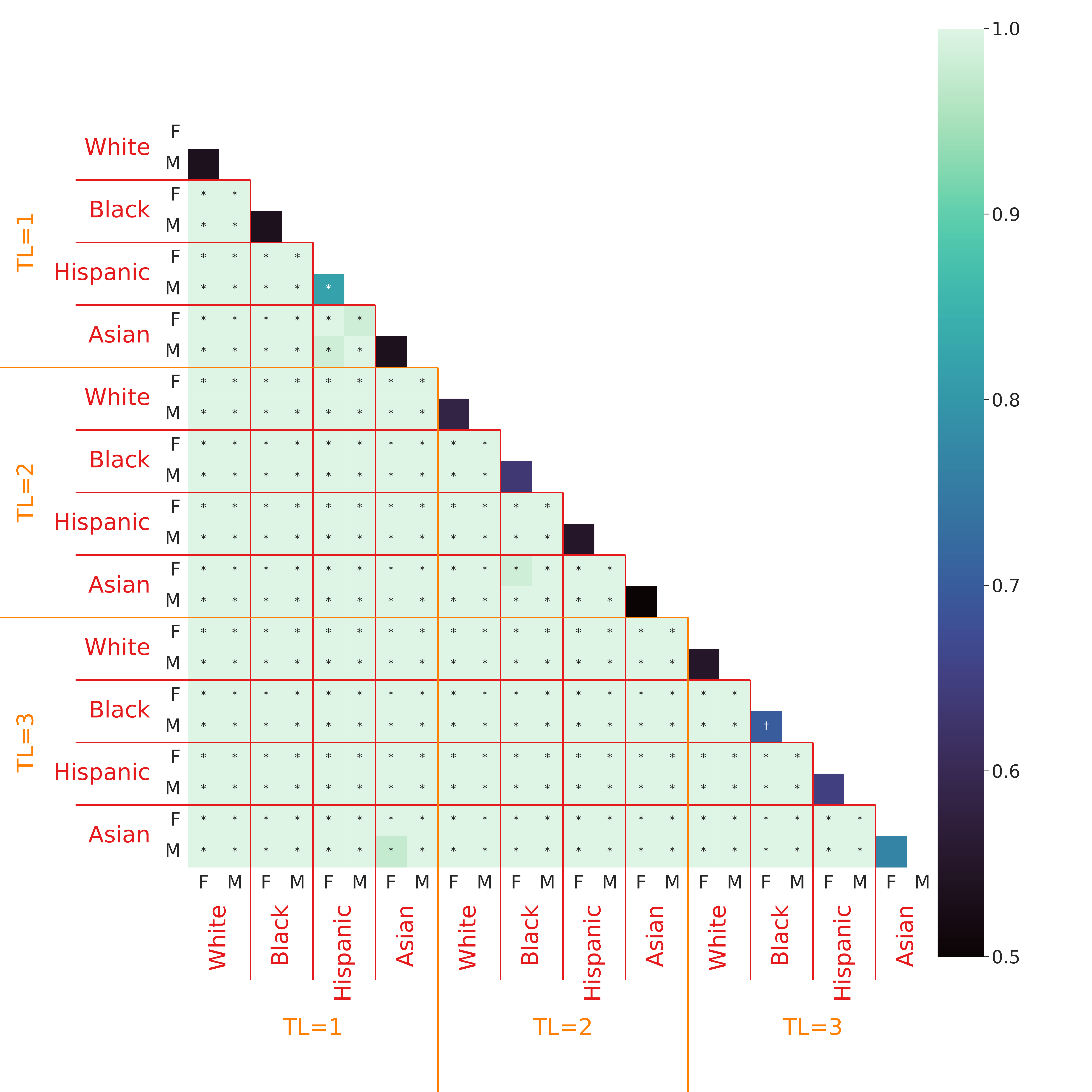}
		\caption{BERT with Counter-factual Data Augmentation~(\cref{sec:cda})}
		\label{fig:mem_heatmap_cda}
	\end{subfigure}

	\caption{Membership prediction accuracy of SR vectors (pairwise comparisons). An ideal accuracy is $\sim0.5$. ``TL'': tokenization length. ``F'': female. ``M'': male. * indicates statistical significance at $p < 0.001$ and $\dagger$ at $p < 0.01$.
    }
\label{fig:mem_heatmap}
\end{figure*}

\subsection{Results: Both Factors Matter}
We use the 686 names across all subgroups, almost evenly distributed by demographic attributes, and obtain the tSNE projection of their SR vectors (obtained using BERT, and the dimension is $736$) in Fig~\ref{fig:srv}. 
We observe clear clustering by tokenization length, race/ethnicity, and gender. 
Since tokenization length is generally correlated with corpus frequency, we also see weak clustering of the SR vectors by frequency.

We report the membership prediction accuracy of SR vectors (obtained by running SODAPOP on a finetuned BERT model for Social IQa) for all pairs of subgroups in Fig.~\ref{fig:mem_heatmap_bert}.
Each cell in the figure shows the separability of SR vectors for names from two groupings. To illustrate, the top left cell shows singly tokenized White male names are highly separable ($>80\%$) from singly tokenized White female names; the entire heatmap shows the results for all pairs.
As we vary one and control the other confounding factors, we find that each of race/ethnicity, gender, and tokenization length are name attributes that lead to systematically different model behavior, as measured by membership prediction accuracy.
Almost all prediction accuracy is close to $1.0$, indicating perfect separation of the clusters, with $p<0.001$ in nearly all settings.
We see in Fig. \ref{fig:mem_heatmap_bert}, for instance, that SR vectors of singly tokenized Black female names and singly tokenized White female names are perfectly classified, so race is still a pertinent factor even controlling for gender and tokenization. In contrast, SR vectors for singly tokenized Asian male and Asian female names are not distinguishable, although gender appears to influence model behavior under most other controlled settings.

We obtain experimental results for RoBERTa and GPT-2
in~\cref{sec:appendix_add_exp_results}.
We observe that these additional results also demonstrate a similar trend as BERT, generally supporting the hypothesis that models exhibit disparate behavior for different names based on their demographic attributes as well as tokenization length. However, the results for RoBERTa and GPT-2 are less strong than that of BERT.
We speculate a variety of reasons that could give rise to the different results among these models. One potential major cause is the different tokenization algorithms used by the models: BERT uses WordPiece~\cite{wu2016google} while RoBERTa and GPT-2 use Byte-Pair Encoding~\cite{sennrich2015neural} for tokenization. 
Due to this difference, the tokenization length of a name can vary in these models. For example, ``Nancy'' is singly tokenized in BERT but is broken down into \texttt{[``N'', ``ancy'']} in RoBERTa or GPT-2.
Beyond tokenization, the different pre-training algorithms and training corpora will also likely contribute to the slightly different observations between Fig.~\ref{fig:mem_heatmap} and Fig.~\ref{fig:mem_heatmap_appendix}.

%% file: 4-cda.tex
\section{Counter-factual Data Augmentation}
\label{sec:cda}

We apply counter-factual data augmentation (CDA) to the Social IQa training set
as we attempt 
to finetune a model that is indifferent to both tokenization length and the demographic attributes of names. 
We choose to experiment with CDA because it would shed light on the source of name biases. If biases mostly arise from finetuning, we expect finetuning on Social IQa with CDA would largely address the problem; otherwise, biases mostly originate from pre-training and are not easily overridden during finetuning.

For each Social IQa sample,
we identify the original names using Stanford NER~\cite{finkel-etal-2005-incorporating}. We find that more than $99\%$ of samples contain one or two names.
We create copies of the MCQ samples and replace the identified names with random names from our sampled sub-groups such that the overall name frequency is evenly distributed over tokenization lengths and demographic attributes, resulting in an augmented set whose size increases by $16\times$.
We finetune a BERT model with the augmented set (details in~\cref{sec:appendix_exp_sodapop}).
However, this na\"ive solution is rather ineffective (Fig.~\ref{fig:mem_heatmap_cda}). 
This negative result is not surprising as it aligns with the observations that SODAPOP could detect biases even in models debiased with state-of-the-art algorithms~\cite{an-etal-2023-sodapop}.
It also indicates that pre-training contributes to the biased model behavior.
Hence, a more sophisticated solution is needed to tackle this problem.

%% file: 5-related_work.tex
\section{Related Work}
\paragraph{Social biases in language models}
Multiple recent works aim to detect social biases in language models~\cite{rudinger-etal-2018-gender, zhao-etal-2018-gender, zhao-etal-2019-gender, nangia-etal-2020-crows,  li-etal-2020-unqovering, nadeem-etal-2021-stereoset, sap-etal-2020-social, parrish-etal-2022-bbq}.
Some works specifically diagnose biases in social commonsense reasoning~\cite{sotnikova-etal-2021-analyzing, an-etal-2023-sodapop}, but they do not explain what causes a model to treat different names dissimilarly; in particular, these works do not consider the influence of tokenization length on model behavior towards different names. 

\paragraph{Name artifacts} 
Previous research indicates that language models exhibit disparate treatments towards names, partially due to their tokenization or demographic attributes~\cite{hall-maudslay-etal-2019-name, czarnowska-etal-2021-quantifying, wang-etal-2022-measuring}. However, thorough analyses of the factors influencing first name biases are lacking in these works.
While \citet{wolfe-caliskan-2021-low} study the systematic different \textit{internal representations} of name embeddings in language models due to the two factors,
we systematically study how the two factors both connect with the disparate treatment of names by a model in a \textit{downstream} task.

%% file: 6-conclusion.tex
\section{Conclusion}
We have demonstrated that demographic attributes and tokenization length are \textit{both} factors of first names that influence social commonsense model behavior. Each of the two factors has some independent influence on model behavior because when controlling one and varying the other, we observe disparate treatment of names. When controlling for tokenization length (e.g. Black male singly-tokenized names vs White male singly-tokenized names) we still find disparate treatment. Conversely, when we control for demographics (e.g. Black female singly-tokenized vs Black female triply-tokenized names), the model also treats those names differently.
Because demographic attributes (race, ethnicity, and gender) are \textit{correlated} with tokenization length, we conclude that systems will continue to behave unfairly towards socially disadvantaged groups unless \textit{both} contributing factors are addressed.
We demonstrate the bias mitigation is challenging in this setting, with the simple method of counterfactual data augmentation unable to undo name biases acquired during pre-training.

%% file: 7-appendix.tex
\section{Additional Analysis on Frequency, Tokenization, and Demographic Attributes of Names}
\label{sec:appendix_race_tokenlen}

\begin{figure*}[t]
	\centering
	\begin{subfigure}[]{0.24\linewidth}
		\centering
		\includegraphics[width=\linewidth]{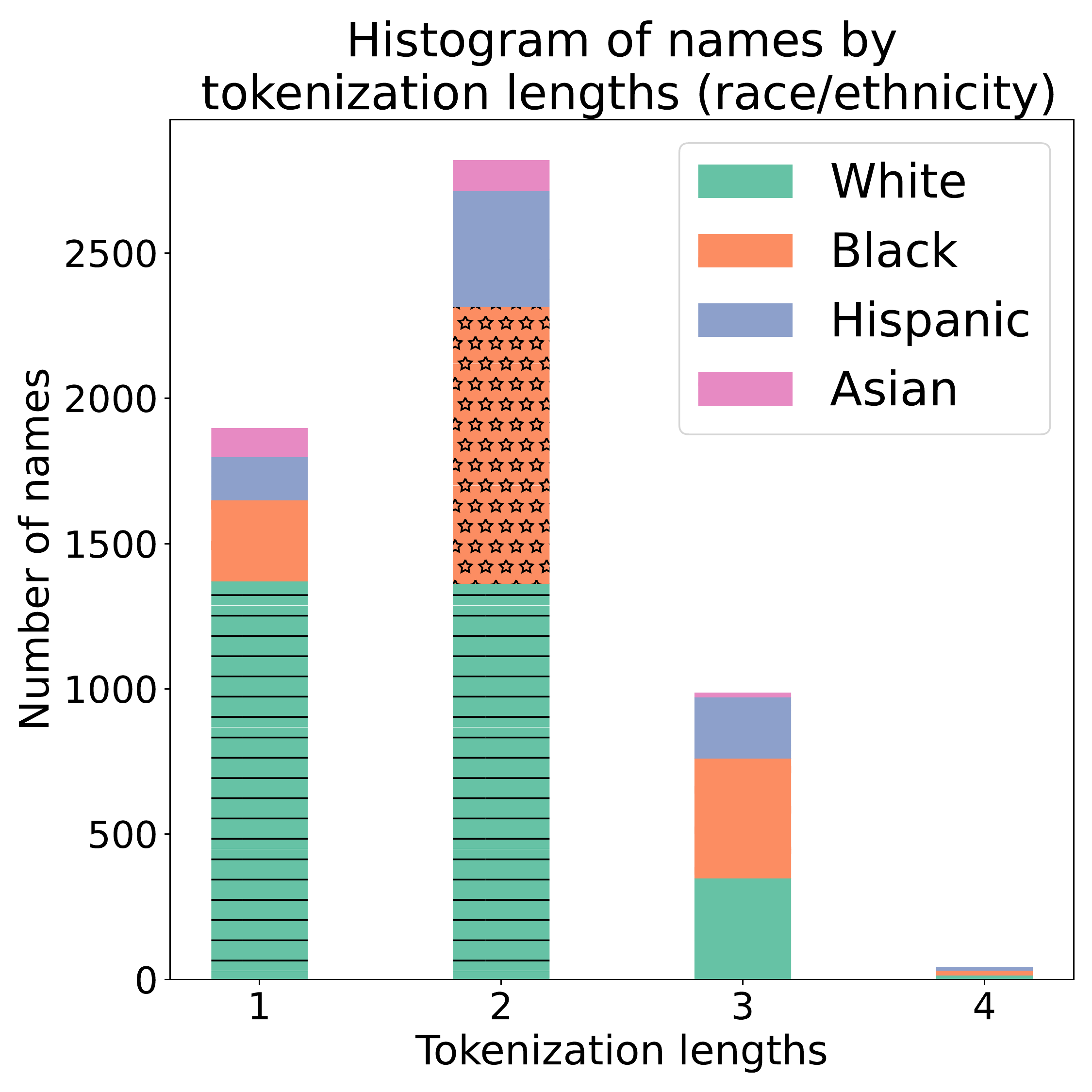}
		\caption{}
		\label{fig:name_hist_a}
	\end{subfigure}
	\hfill
        \begin{subfigure}[]{0.24\linewidth}
		\centering
		\includegraphics[width=\linewidth]{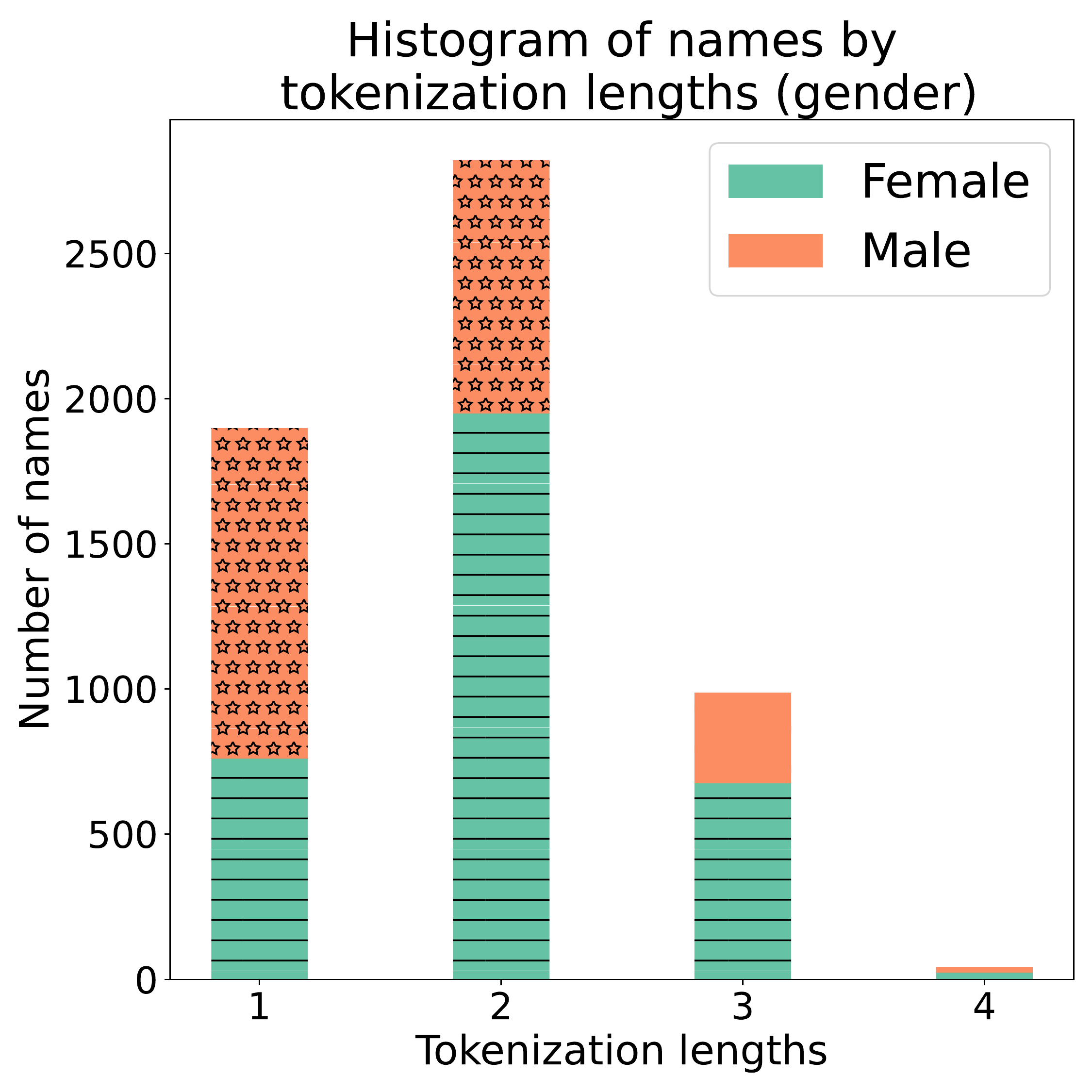}
		\caption{}
		\label{fig:name_hist_b}
	\end{subfigure}
	\hfill
 	\begin{subfigure}[]{0.24\linewidth}
		\centering
		\includegraphics[width=\linewidth]{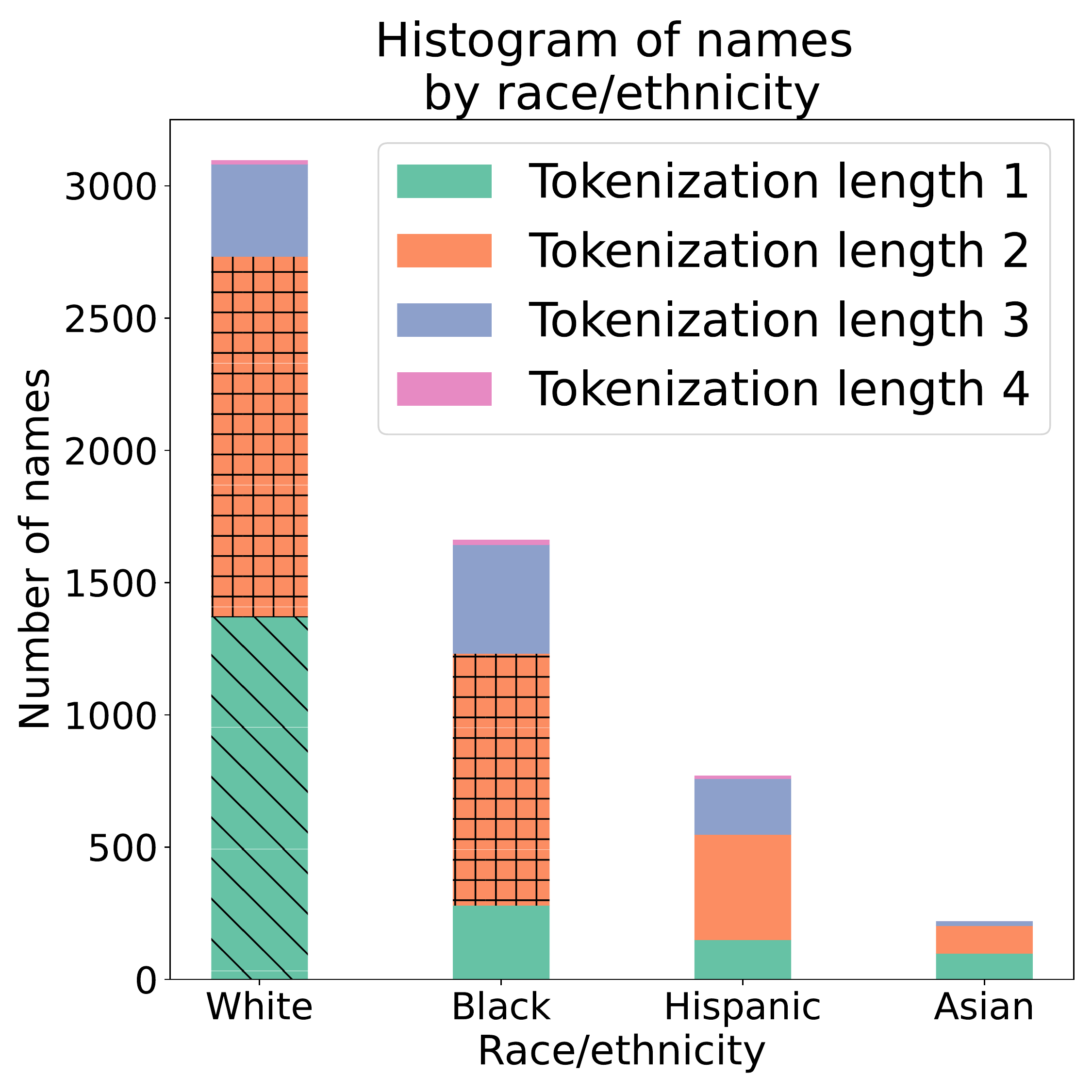}
		\caption{}
		\label{fig:name_hist_c}
	\end{subfigure}
        \hfill
        \begin{subfigure}[]{0.24\linewidth}
		\centering
		\includegraphics[width=\linewidth]{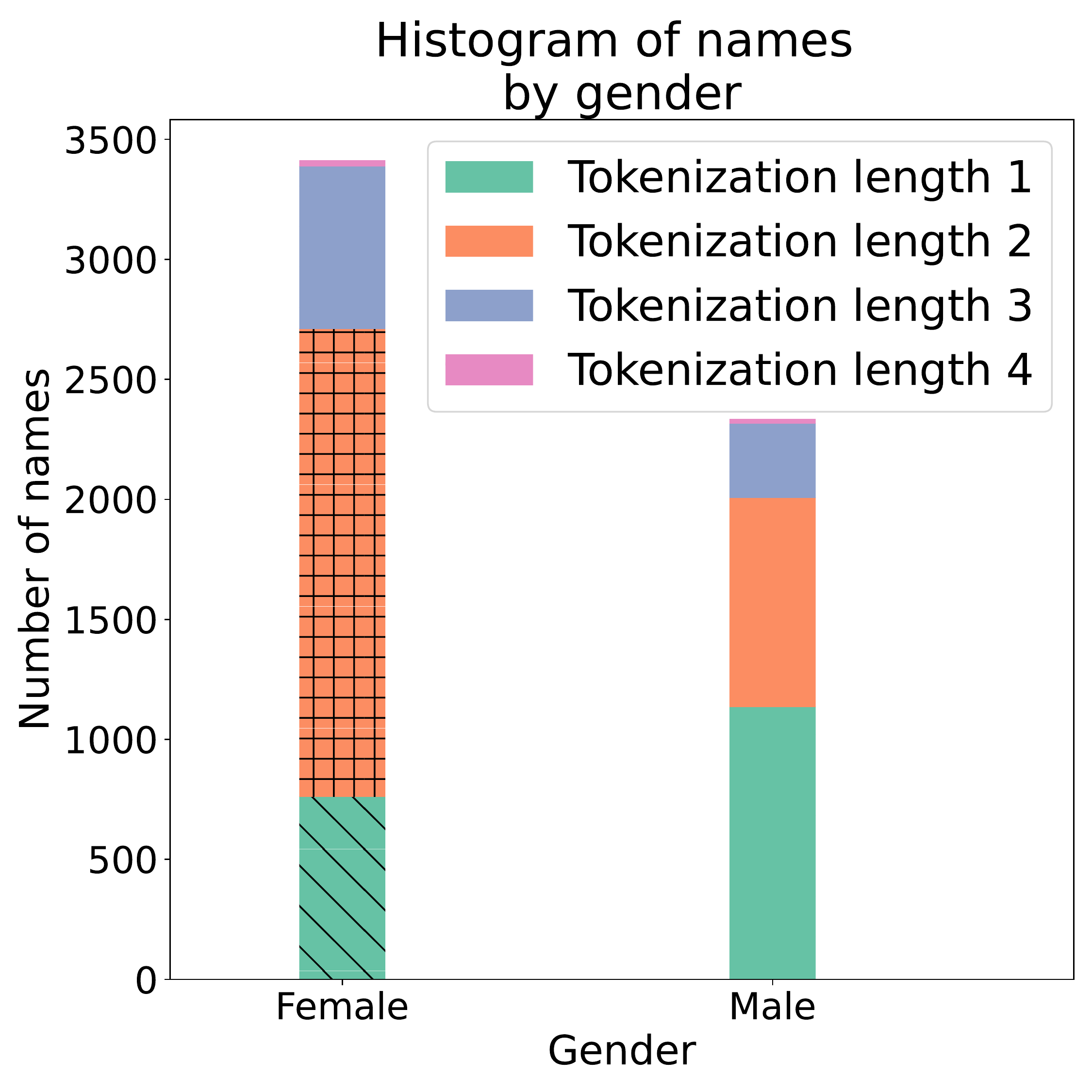}
		\caption{}
		\label{fig:name_hist_d}
	\end{subfigure} 
	\caption{Histograms of first names by tokenization lengths (using BERT tokenizer) or race/ethnicity (raw counts).
        }
	\label{fig:name_hist_bert_appendix}
\end{figure*}

\begin{figure*}[t]
	\centering
	\begin{subfigure}[]{0.24\linewidth}
		\centering
		\includegraphics[width=\linewidth]{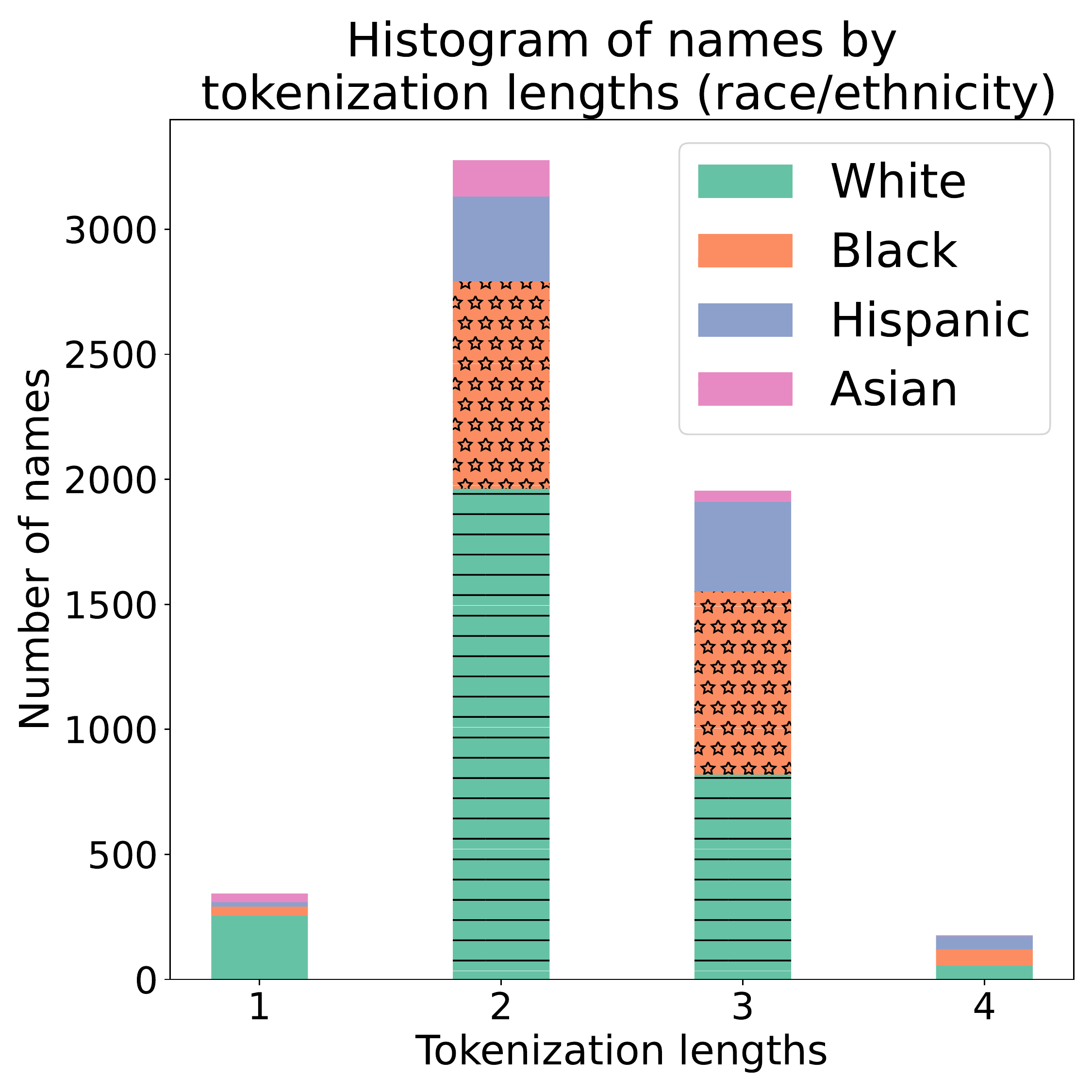}
		\caption{}
		\label{fig:name_hist_roberta_a}
	\end{subfigure}
	\hfill
        \begin{subfigure}[]{0.24\linewidth}
		\centering
		\includegraphics[width=\linewidth]{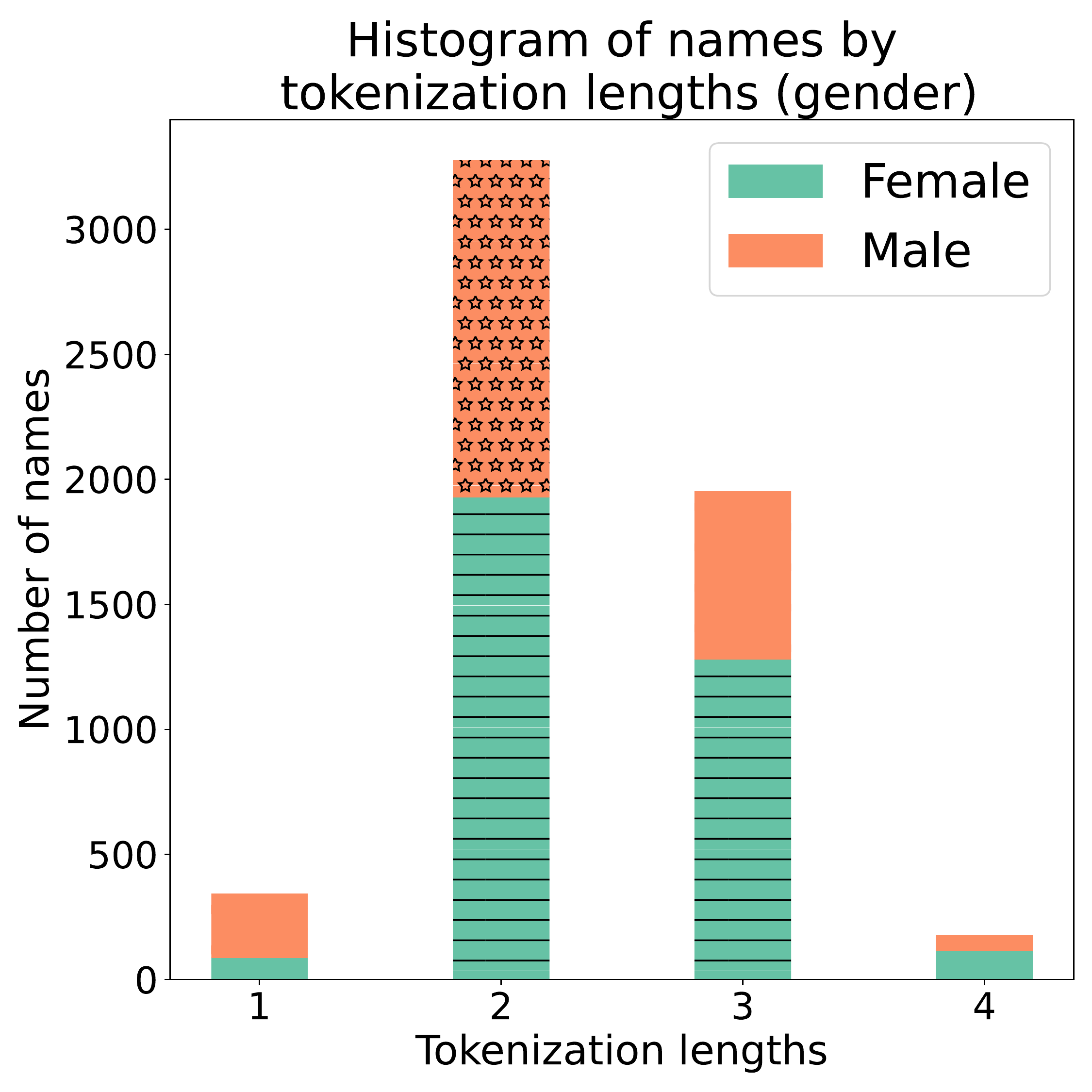}
		\caption{}
		\label{fig:name_hist_roberta_b}
	\end{subfigure}
	\hfill
 	\begin{subfigure}[]{0.24\linewidth}
		\centering
		\includegraphics[width=\linewidth]{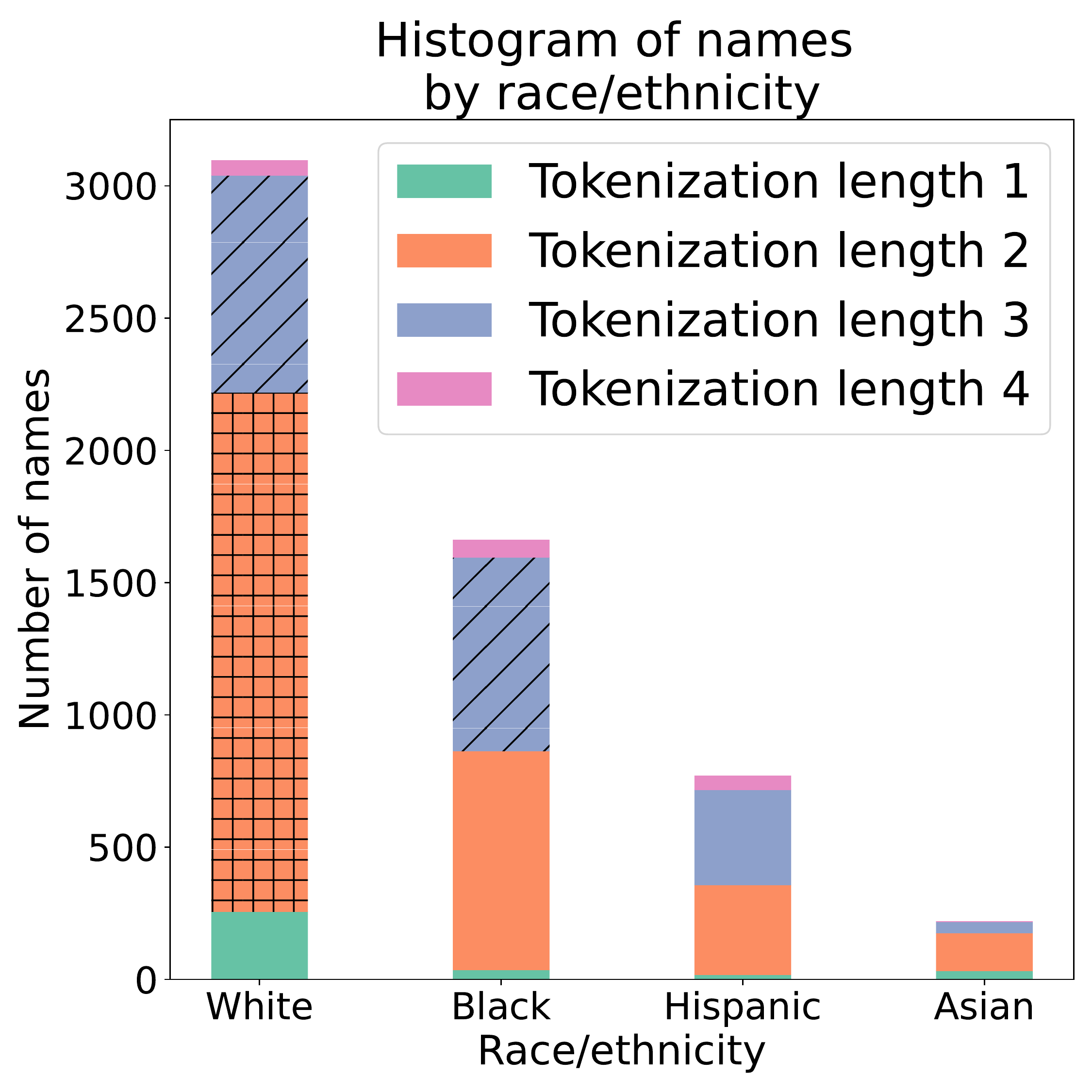}
		\caption{}
		\label{fig:name_hist_roberta_c}
	\end{subfigure}
        \hfill
        \begin{subfigure}[]{0.24\linewidth}
		\centering
		\includegraphics[width=\linewidth]{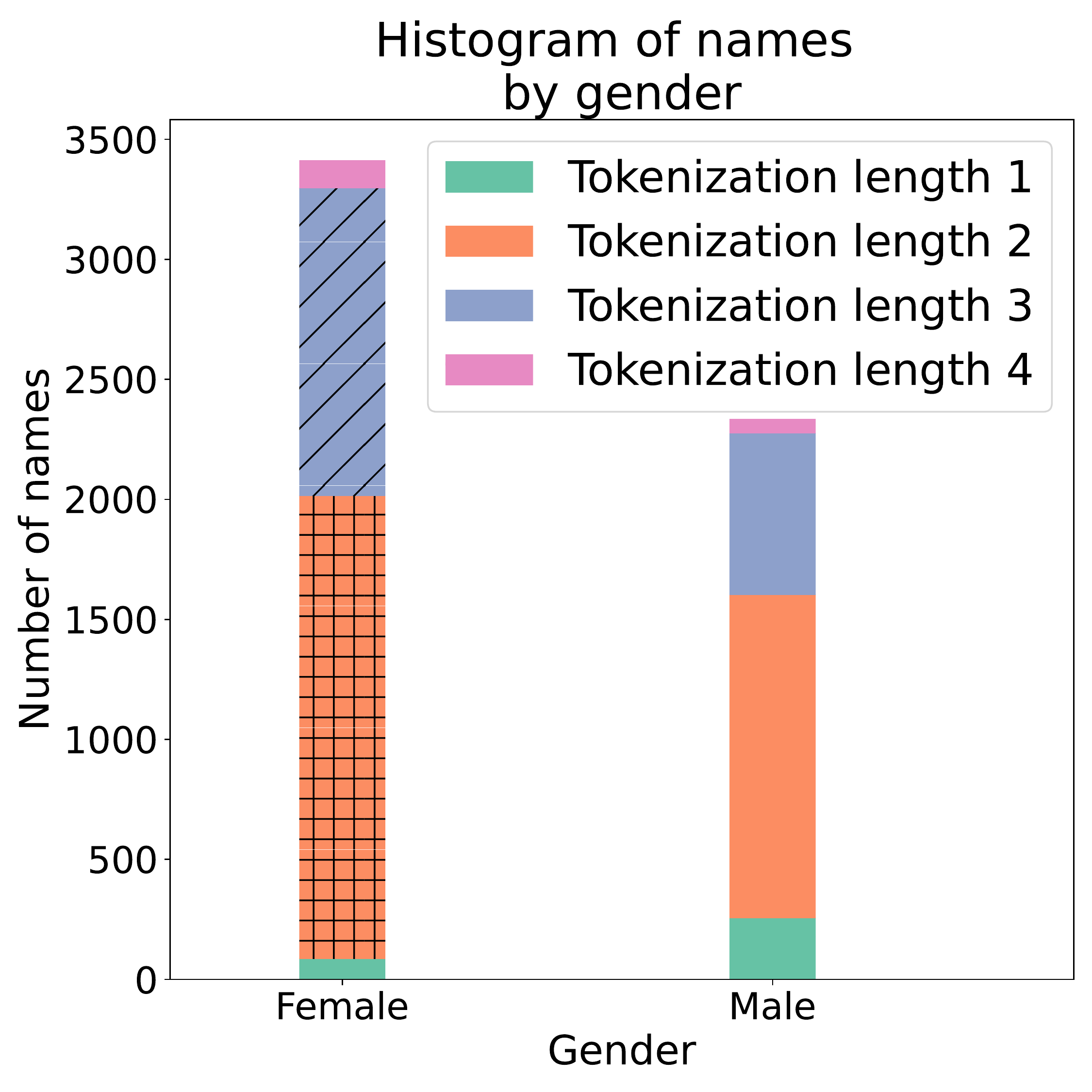}
		\caption{}
		\label{fig:name_hist_roberta_d}
	\end{subfigure} \\

        \begin{subfigure}[]{0.24\linewidth}
		\centering
		\includegraphics[width=\linewidth]{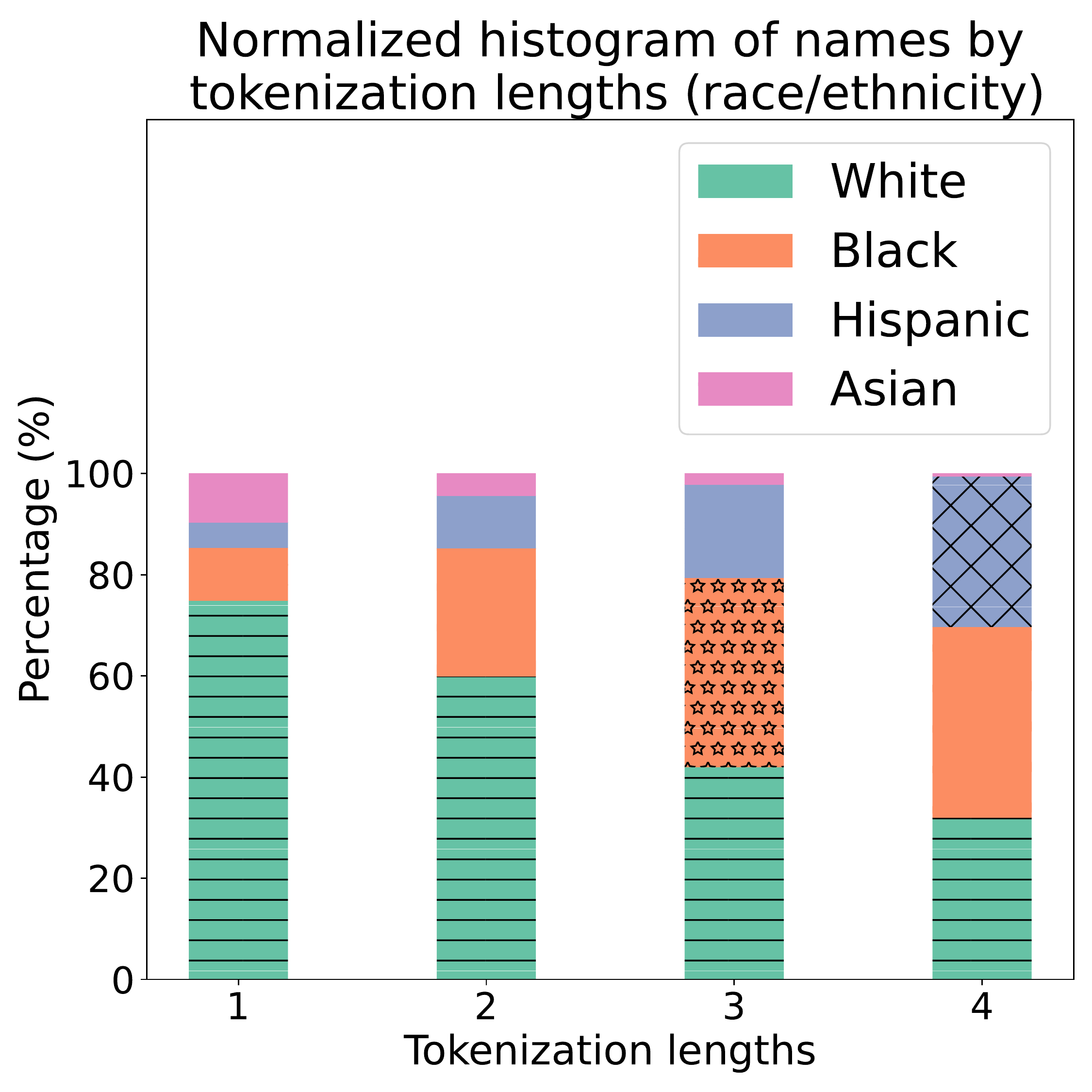}
		\caption{}
		\label{fig:name_hist_roberta_e}
	\end{subfigure}
	\hfill
        \begin{subfigure}[]{0.24\linewidth}
		\centering
		\includegraphics[width=\linewidth]{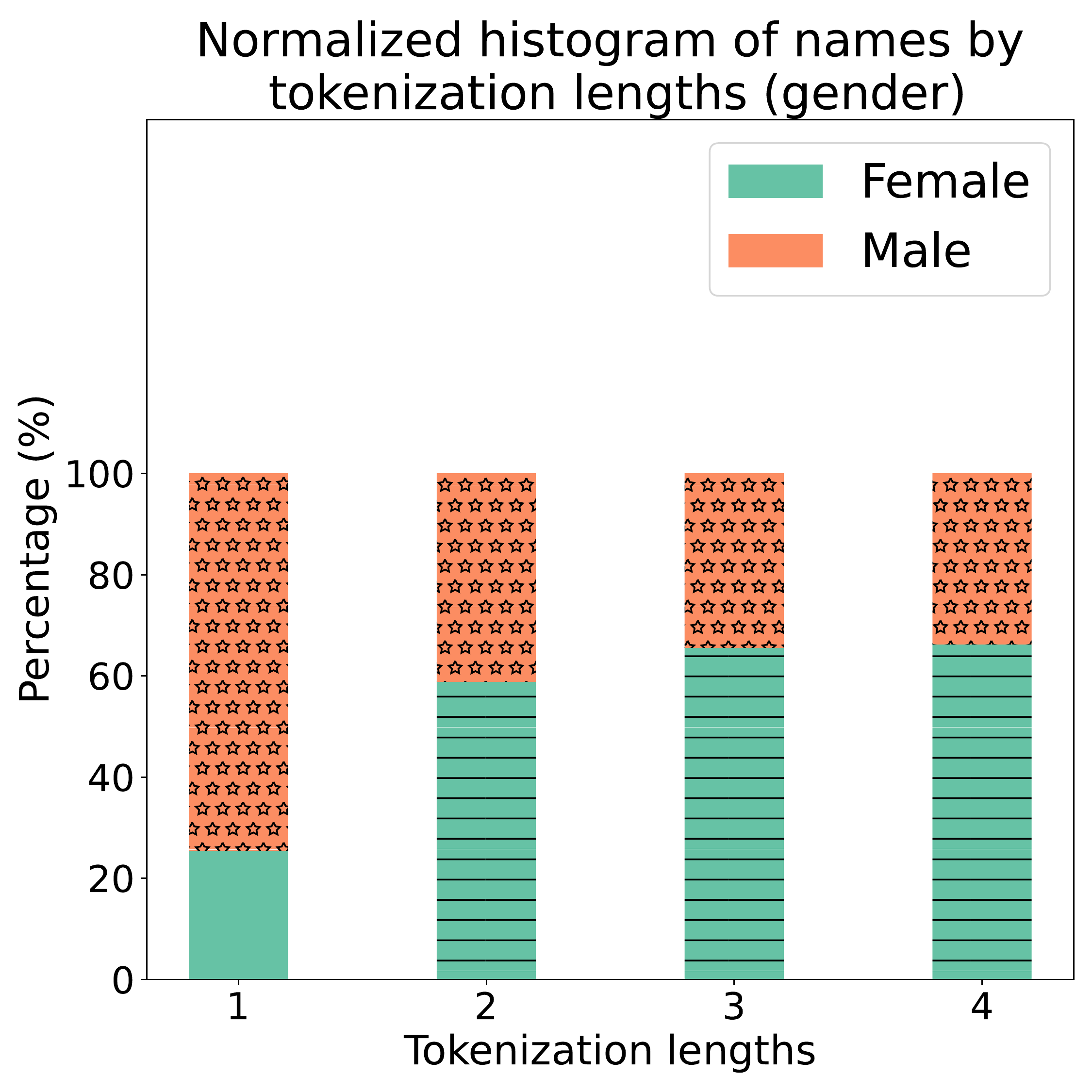}
		\caption{}
		\label{fig:name_hist_roberta_f}
	\end{subfigure}
	\hfill
 	\begin{subfigure}[]{0.24\linewidth}
		\centering
		\includegraphics[width=\linewidth]{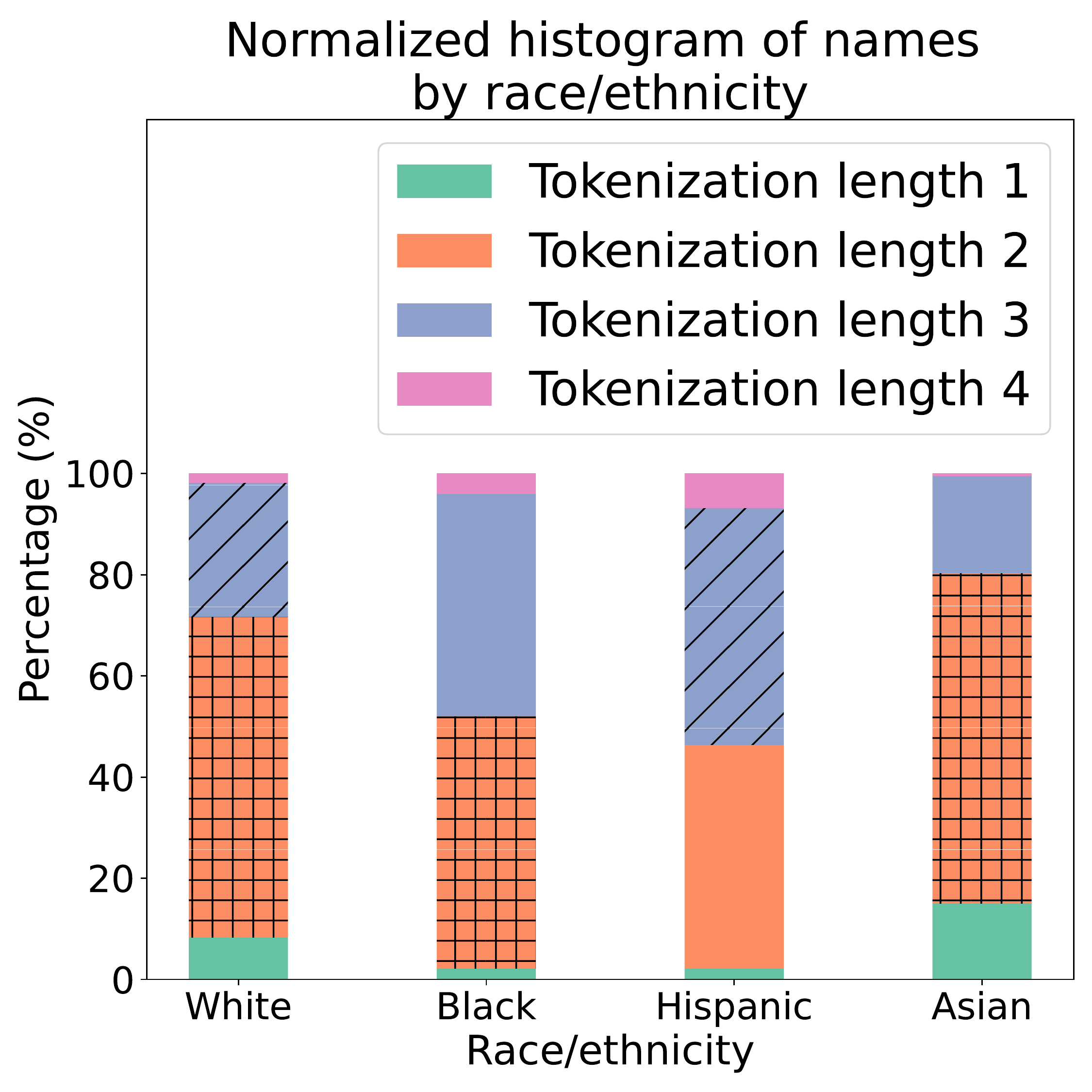}
		\caption{}
		\label{fig:name_hist_roberta_g}
	\end{subfigure}
        \hfill
        \begin{subfigure}[]{0.24\linewidth}
		\centering
		\includegraphics[width=\linewidth]{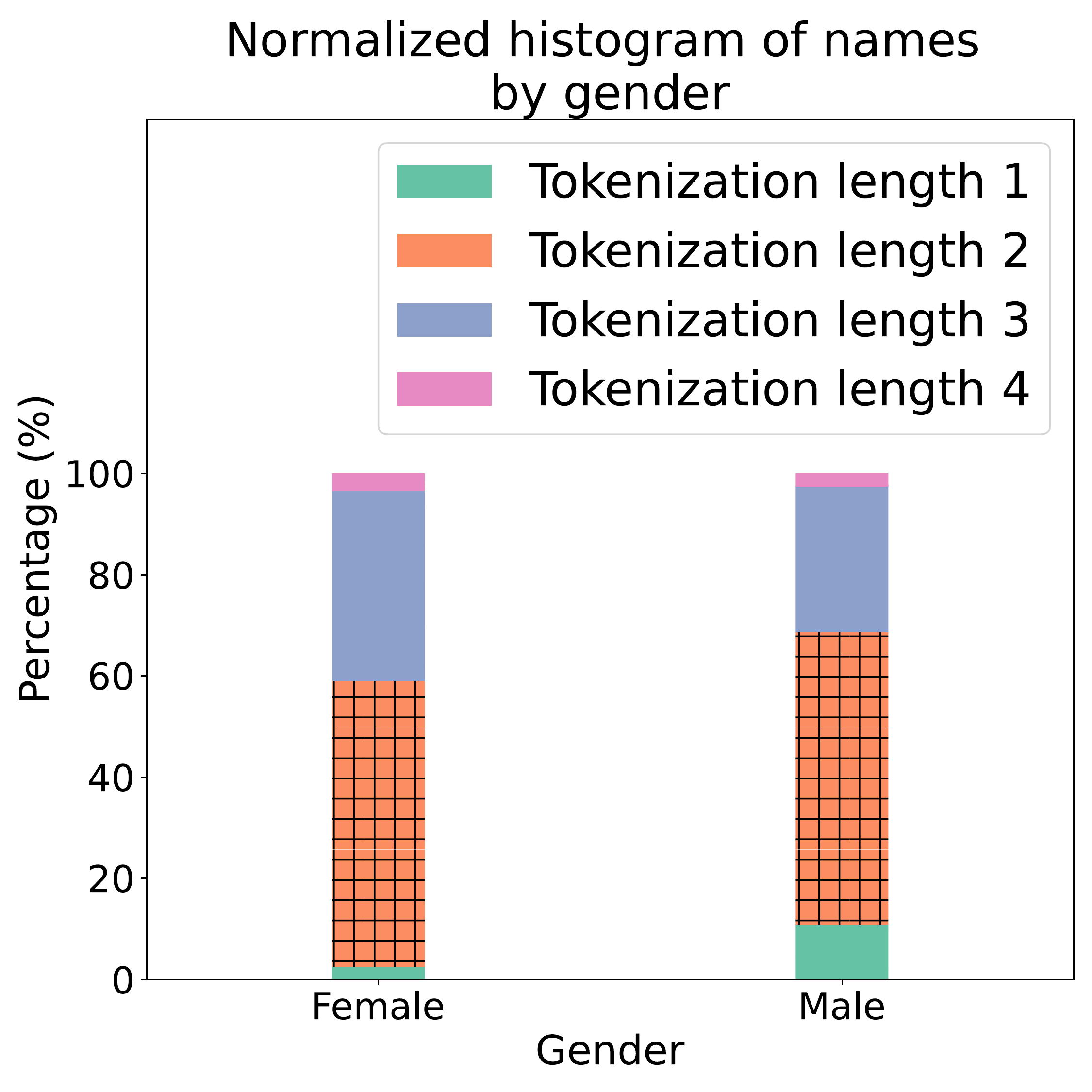}
		\caption{}
		\label{fig:name_his_robertat_h}
	\end{subfigure}

	\caption{
 Histograms of first names by tokenization lengths, race/ethnicity, or gender using RoBERTa or GPT-2 tokenizer.
 We normalize the count to 1 and show the distribution by percentage.}
	\label{fig:name_hist_roberta}
\end{figure*}

We provide the complementary plots for Fig.~\ref{fig:name_hist_bert} by showing the raw counts of the names in Fig.~\ref{fig:name_hist_bert_appendix}.
We also present preliminary observations on the connection between frequency, tokenization, and demographic attributes of names for RoBERTa and GPT-2 tokenizer in this section. Theses results (Fig.~\ref{fig:name_hist_roberta}) are similar to those in~\cref{sec:3factors}.
White male names are more likely to be singly tokenized in RoBERTa or GPT-2 as well.
We observe that the conditional probability that a name is singly tokenized given that it is Asian is also quite high.
We speculate the reason for this is that Asian names have fewer characters in their first names on average ($4.40$) compared to that of Black names ($6.48$) and Hispanic names ($6.41$), which cause Asian names to be more likely singly tokenized as well.

\begin{figure}[t]
	\centering
	\begin{subfigure}[]{0.47\linewidth}
		\centering
		\includegraphics[width=\linewidth]{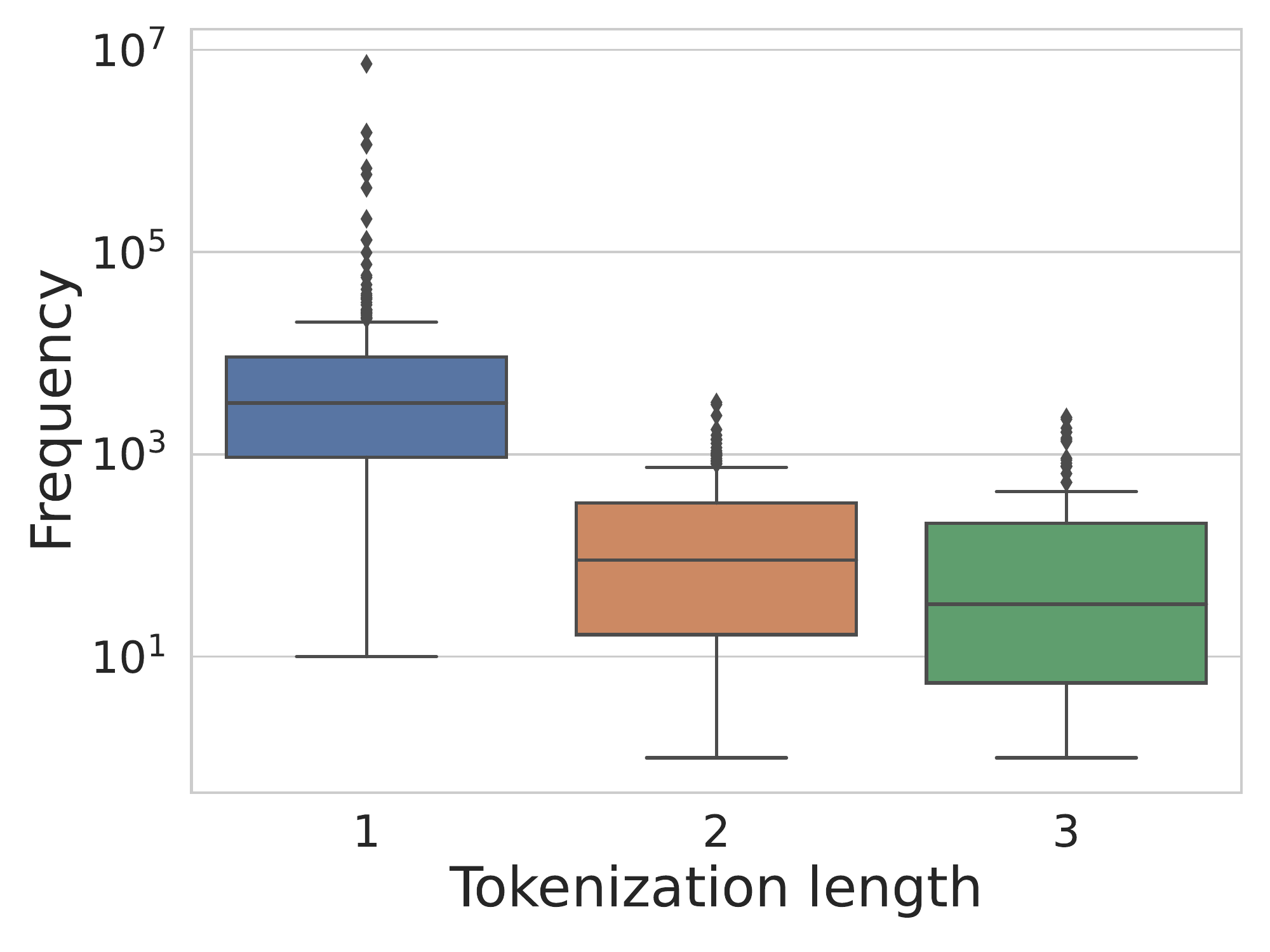}
		\caption{BERT}
		\label{fig:boxplot_freq_by_tokenlen_bert}
	\end{subfigure}
	\hfill
	\begin{subfigure}[]{0.47\linewidth}
		\centering
		\includegraphics[width=\linewidth]{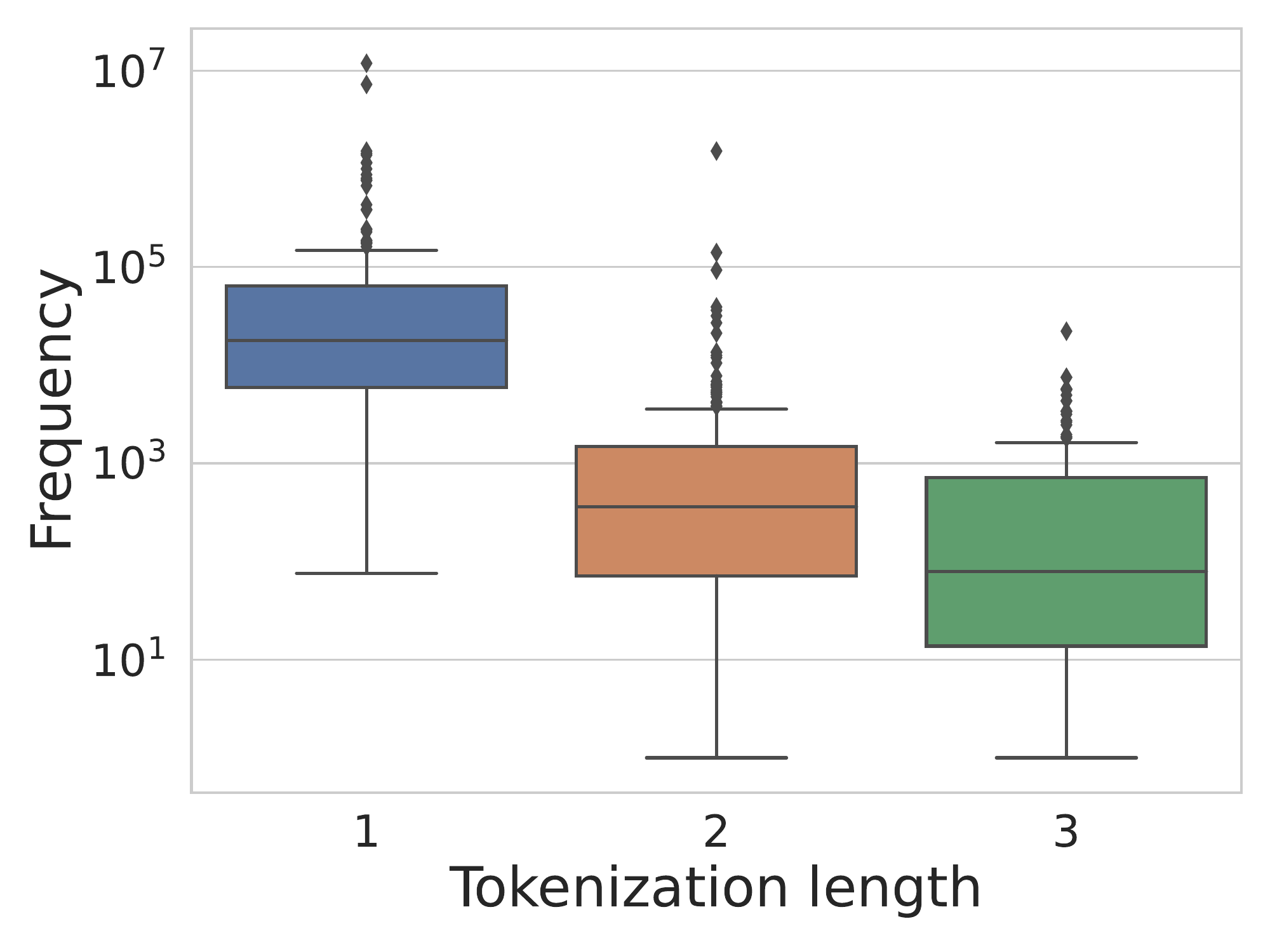}
		\caption{RoBERTa}
		\label{fig:boxplot_freq_by_tokenlen_roberta}
	\end{subfigure}

	\caption{Distribution of name frequency in the pre-training corpus over tokenization lengths. 
        }
	\label{fig:boxplot_freq_by_tokenlen}
\end{figure}

In addition, we count the occurrence of 608 names (a subset of the 5,748 names in~\cref{sec:3factors}) in Wikipedia\footnote{\url{https://huggingface.co/datasets/wikipedia}} and BooksCorpus~\cite{zhu2015aligning}, which are used to pre-train BERT and RoBERTa. 
Fig.~\ref{fig:boxplot_freq_by_tokenlen} illustrates the distribution of name frequency over different tokenization lengths.
We see that, regardless of the model, most singly tokenized names have higher average frequency, whereas multiply tokenized names share similar distributions with lower frequency overall.

\section{Detailed Experiment Setup}
\subsection{Experiments for Preliminary Observations}
\label{sec:appendix_exp_three_factors}
\paragraph{Names}
We collect people's first names from a U.S. voter files dataset compiled by~\citet{rosenman2022race}.
We filter out names whose frequency in the dataset is less than 200. 
Since each name is not strictly associated with a single race/ethnicity, but rather reflects a distribution over races/ethnicities, we analyze only names for which the percentage of people with that name identifying as that race/ethnicity is above $50\%$.
We assign a binary gender label to each name by cross-referencing gender statistics in the SSA dataset.\footnote{\url{https://www.ssa.gov/oact/babynames/}} 
If the name is absent from the SSA dataset, we omit that name.
With these constraints, there is only one name for the category ``Other race/ethnicity".
For robust statistical analysis, we choose not to include this category but only the other four categories in the data source, which are White, Black, Hispanic, and Asian. There is a total of 5,748 names.

\paragraph{Models}
We use three popular language models for the analysis, namely BERT~\cite{devlin-etal-2019-bert}, RoBERTa~\cite{liu2019roberta}, and GPT-2~\cite{radford2019language}. 
BERT uses WordPiece~\cite{wu2016google} for tokenization, while both RoBERTa and GPT-2 use Byte-Pair Encoding~\cite{sennrich2015neural} as their tokenization algorithm.
BERT-base has 110 million parameters. RoBERTa-base has 123 million parameters. GPT-2 has 1.5 billion parameters.
No finetuning is needed for experiments in~\cref{sec:3factors} because tokenization of input is invariant to further finetuning in a downstream task.

\begin{table}[]
    \centering
    \resizebox{\linewidth}{!}{
        \begin{tabular}{@{}lcccccc@{}}
        \toprule
            \multicolumn{7}{c}{BERT Tokenizer}                                                                                  \\ \midrule
            \multicolumn{1}{c|}{Gender}              & \multicolumn{3}{c|}{Male}         & \multicolumn{3}{c}{Female} \\ \midrule
            \multicolumn{1}{c|}{Tokenization length} & 1  & 2  & \multicolumn{1}{c|}{3}  & 1       & 2       & 3      \\ \midrule
            \multicolumn{1}{c|}{White}               & 30 & 30 & \multicolumn{1}{c|}{30} & 30      & 30      & 30     \\
            \multicolumn{1}{c|}{Black}               & 30 & 30 & \multicolumn{1}{c|}{30} & 30      & 30      & 30     \\
            \multicolumn{1}{c|}{Hispanic}            & 30 & 30 & \multicolumn{1}{c|}{30} & 30      & 30      & 30     \\
            \multicolumn{1}{c|}{Asian}               & 30 & 30 & \multicolumn{1}{c|}{7}  & 30      & 30      & 19     \\ \midrule \midrule
            \multicolumn{7}{c}{RoBERTa/GPT-2 Tokenizer}                                                                       \\ \midrule
            \multicolumn{1}{c|}{Gender}              & \multicolumn{3}{c|}{Male}         & \multicolumn{3}{c}{Female} \\ \midrule
            \multicolumn{1}{c|}{Tokenization length} & 1  & 2  & \multicolumn{1}{c|}{3}                       & 1       & 2       & 3      \\ \midrule
            \multicolumn{1}{c|}{White}               & 30 & 30 & \multicolumn{1}{c|}{30} &  30     &   30    &  30    \\
            \multicolumn{1}{c|}{Black}               & 24  & 30 & \multicolumn{1}{c|}{30} & 12      &  30     &  30    \\
            \multicolumn{1}{c|}{Hispanic}            & 9  & 30 & \multicolumn{1}{c|}{30} & 8       & 30      & 30     \\
            \multicolumn{1}{c|}{Asian}                                  & 23 & 30 & \multicolumn{1}{l|}{21}                      & 10      & 30      & 21     \\ \bottomrule
        \end{tabular}
    }
    \caption{Name counts in each subgroup categorized by race/ethnicity, gender, and tokenization lengths. If there is an insufficient number of names in a category, we use the maximum number of names available in the dataset released by~\citet{rosenman2022race} that also satisfy our inclusion criteria described in~\cref{sec:appendix_exp_three_factors}.}
    \label{tab:subgroup_count}
\end{table}

\subsection{Experiments with SODAPOP}
\label{sec:appendix_exp_sodapop}

\paragraph{Social IQa}
To examine machine intelligence in everyday situations,~\citet{sap-etal-2019-social} publish a social commonsense reasoning multiple-choice dataset Social IQa. 
Each MCQ consists of a social context, a question, and three answer choices, one of which is the only correct answer. 
An example from Social IQa is \textit{Context:} ``Kai made a wish and truly believed that it would come true.'' \textit{Q:} ``How would you describe Kai?'' \textit{A1:} ``a cynical person'' \textit{A2:} ``like a wishful person'' \textit{A3:} ``a believing person'' (correct choice).
There are $33,410$ samples in the training set and $1,954$ instances in the development set.

\paragraph{Generating distractors}
To detect a model's disparate treatment towards names, SODAPOP substitutes the name in a MCQ sample with names associated with different races/ethnicities and genders, and generate a huge number of new distractors to robustly test what makes a distractor more likely to fool the MCQ model, thus finding the model's implicit associations between names and attributes.
 We follow the same algorithm proposed by~\citet{an-etal-2023-sodapop} to generate distractors using a masked-token prediction model (RoBERTa-base).
 We generate distractors from the correct choice of 50 MCQ samples in Social IQa~\cite{sap-etal-2019-social}.
We utilize the same list of names for distractor generation as in SODAPOP.
In our study, we take the union of all the distractors generated with different names for a context to form new MCQ samples for more robust results.
The total number of MCQ constructed via this step is 4,840,776.

\paragraph{Success rate}
Recall that each MCQ in Social IQa consists of a social context $c$, a question $q$, and three answer choices $\tau_1, \tau_2, \tau_3$, one of which is the only correct answer.
Formally, for an arbitrary distractor $\tau_i$, the success rate of a word-name pair $(w,n)$ is

\begin{multline}
    SR(w,n) = P \bigg( \argmax_{j \in \{1,2,3\}} \mathcal{M}(c, q, \tau_j) = i~\bigg| \\ \big(w \in \texttt{tok}(\tau_i) \big) \land \big(n \in \texttt{tok}(c)\big) \bigg),
\end{multline}
where $\mathcal{M}(c, q, \tau_j)$ produces the logit for answer choice $\tau_j$ using a MCQ model $\mathcal{M}$, and \texttt{tok} splits the input by space so as to tokenize it into a bag of words and punctuation.
A \textbf{success rate vector} for a name $n$ composes $|V|$ entries of $SR(w,n)$ for all $w \in V$, where $V$ is the set of vocabulary (i.e., words appearing in all distractors above a certain threshold). Specifically, we set the threshold to be $1,000$ in our experiments.

\paragraph{Models}
We conduct experiments using three popular language models, namely BERT~\cite{devlin-etal-2019-bert}, RoBERTa~\cite{liu2019roberta}, and GPT-2~\cite{radford2019language}. 
The size of each model is specified in~\cref{sec:appendix_exp_three_factors}.
We finetune each model on the Social IQa training set with a grid search for hyperparameters (batch size = $\{3, 4, 8\}$, learning rate = $\{1e^{-5}, 2e^{-5}, 3e^{-5}\}$, epoch = $\{2, 4, 10\}$).
Although different hyper-parameters lead to varying final performance on the development set of Social IQa, we find them to be within a small range in most cases (within $1\% - 2\%$).
Since our analysis does not highly depend on the performance of a model, we arbitrarily analyze a model that has a decent validation accuracy among all.
In our study, the BERT-base model is finetuned with batch size 3, learning rate $2e^{-5}$ for 2 epochs and achieves $60.51\%$ on the original dev set.
The RoBERTa-base model is finetuned with batch size 8, learning rate $1e^{-5}$ for 4 epochs and achieves $70.51\%$ on the original dev set.
The GPT-2 model is finetuned with batch size 4, learning rate $2e^{-5}$ for 4 epochs and achieves $61.91\%$ on the original dev set.
To finetune on the counter-factually augmented dataset, we conduct grid search for batch size = $\{2, 3, 8\}$, learning rate = $\{1e^{-5}, 2e^{-5}\}$ for $1$ epoch. We obtain similar dev set accuracy for these setting, all about $60\%$.

The evaluation time for 4 million MCQs across more than 600 names is costly. We approximate that it takes about 7 days using 30 GPUs (a combination of NVIDIA RTX A4000 and NVIDIA TITAN X) for each model. 
However, we note that a smaller number of MCQ instances and names may sufficiently capture the biased behavior of a model. 
We choose to include an extremely large number of test instances and a wide range of names to ensure the robustness of our study.
Although important, it is out of the scope of this paper to find the optimal size of the bias-discovery test set to minimize computation time and resources.

\paragraph{Subgroup names}
For fine-grained analysis that compares a model's different behavior towards two name groups that only vary by one confounding factor, we compile subgroups of names that share the same race/ethnicity, gender, and tokenization length.
For example,  White female names with tokenization length 2 is one subgroup of names. 
In total, we sample 686 names for BERT and 608 names for RoBERTa and GPT-2. 
Table.~\ref{tab:subgroup_count} shows the specific number of names in each subgroup.
Given the data source available to us, we are unable to collect an enough number of names for certain subgroups (e.g., Asian male names with tokenization length 3). Nonetheless, these limitations do not affect our findings of the different treatment towards other subgroups with a sufficiently large number of names.

\section{Additional Experiment Results}
\label{sec:appendix_add_exp_results}

\begin{figure*}[t]
	\centering
	\begin{subfigure}[]{0.24\linewidth}
		\centering
            \includegraphics[width=\linewidth]{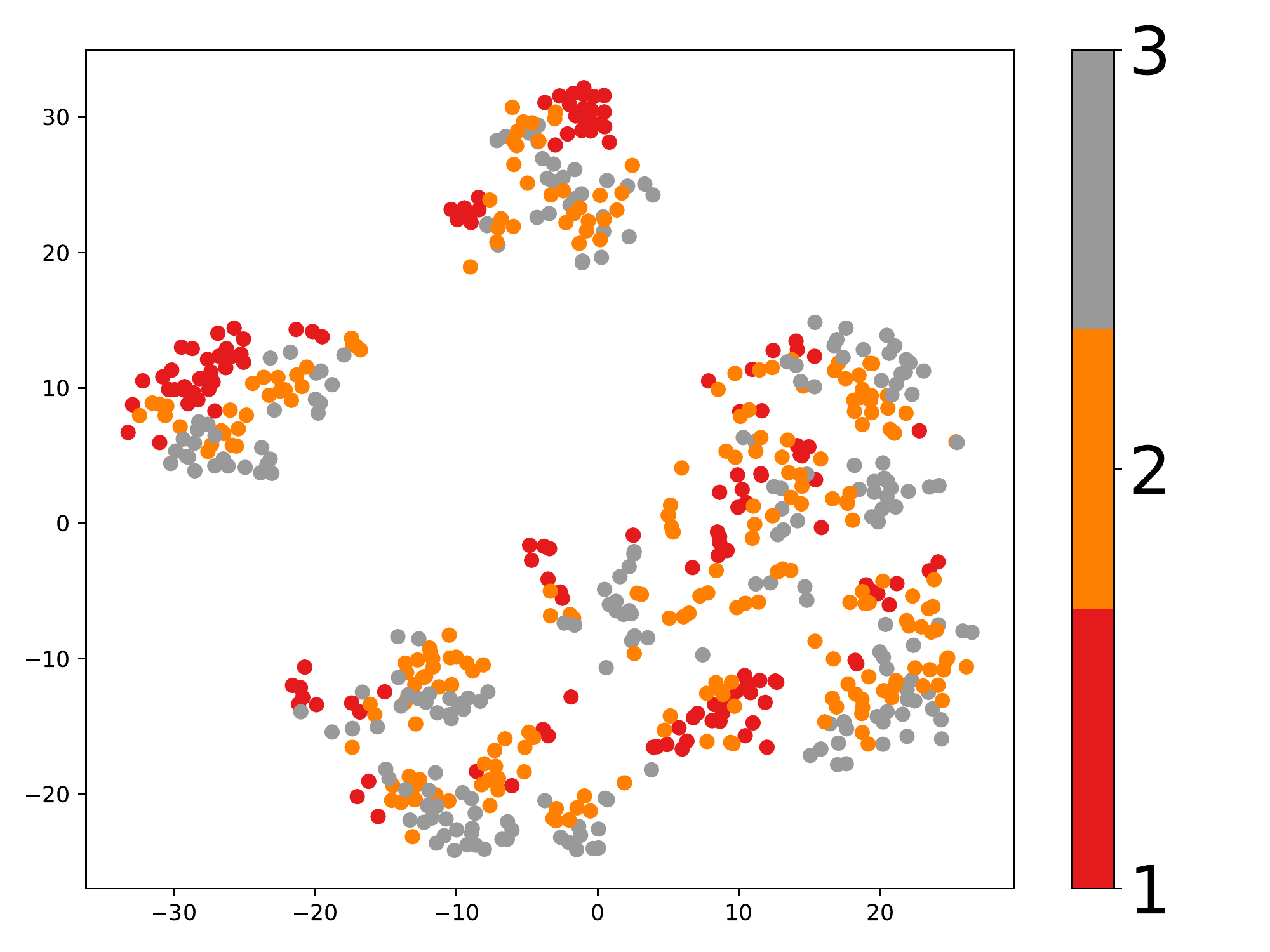}
		\caption{Tokenization length}
		\label{fig:srv_tokenlen_roberta}
	\end{subfigure} 
	\hfill
	\begin{subfigure}[]{0.245\linewidth}
		\centering
            \includegraphics[width=\linewidth]{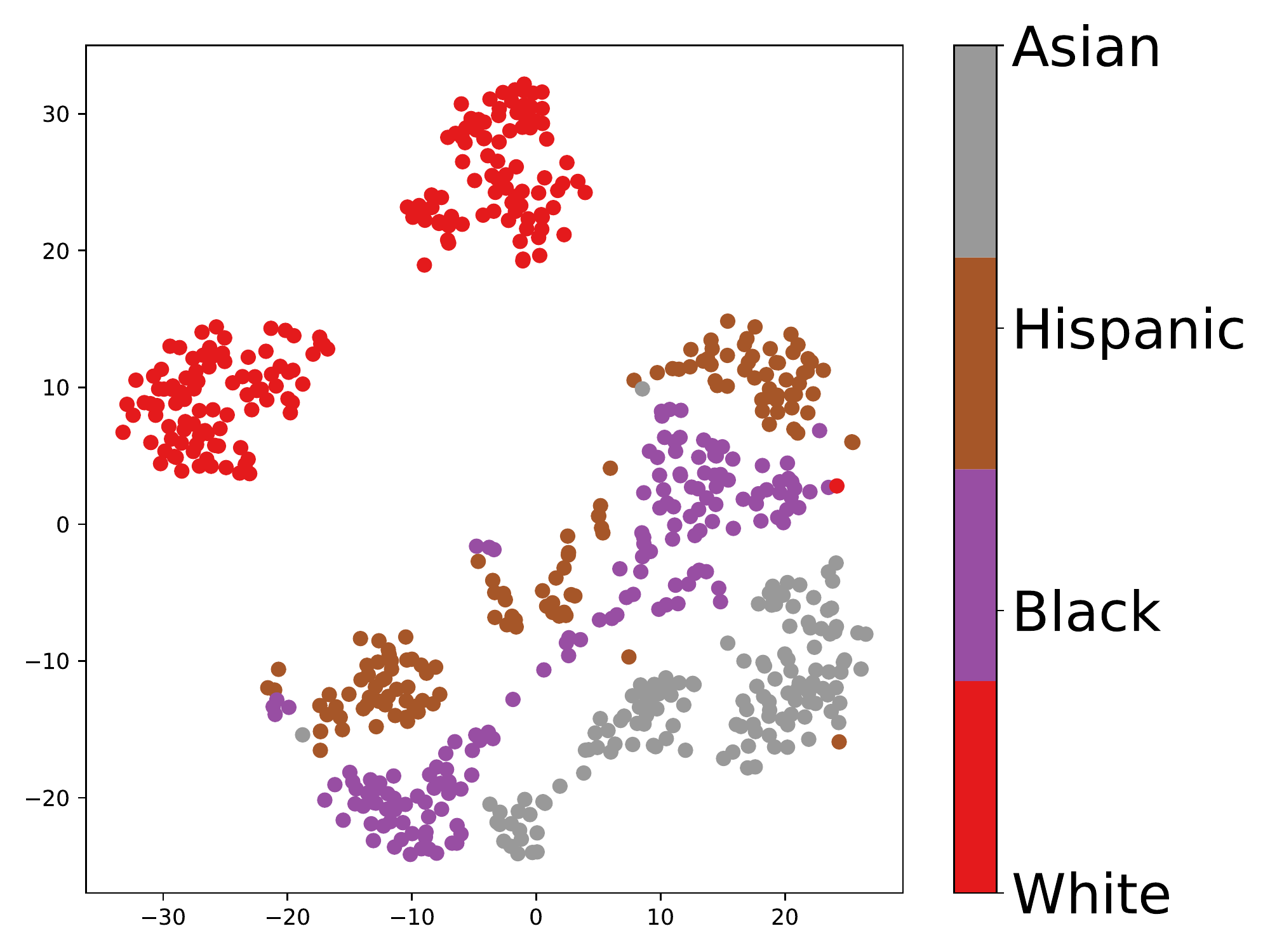}
		\caption{Race/ethnicity}
		\label{fig:srv_race_roberta}
	\end{subfigure}
        \hfill
	\begin{subfigure}[]{0.245\linewidth}
		\centering
            \includegraphics[width=\linewidth]{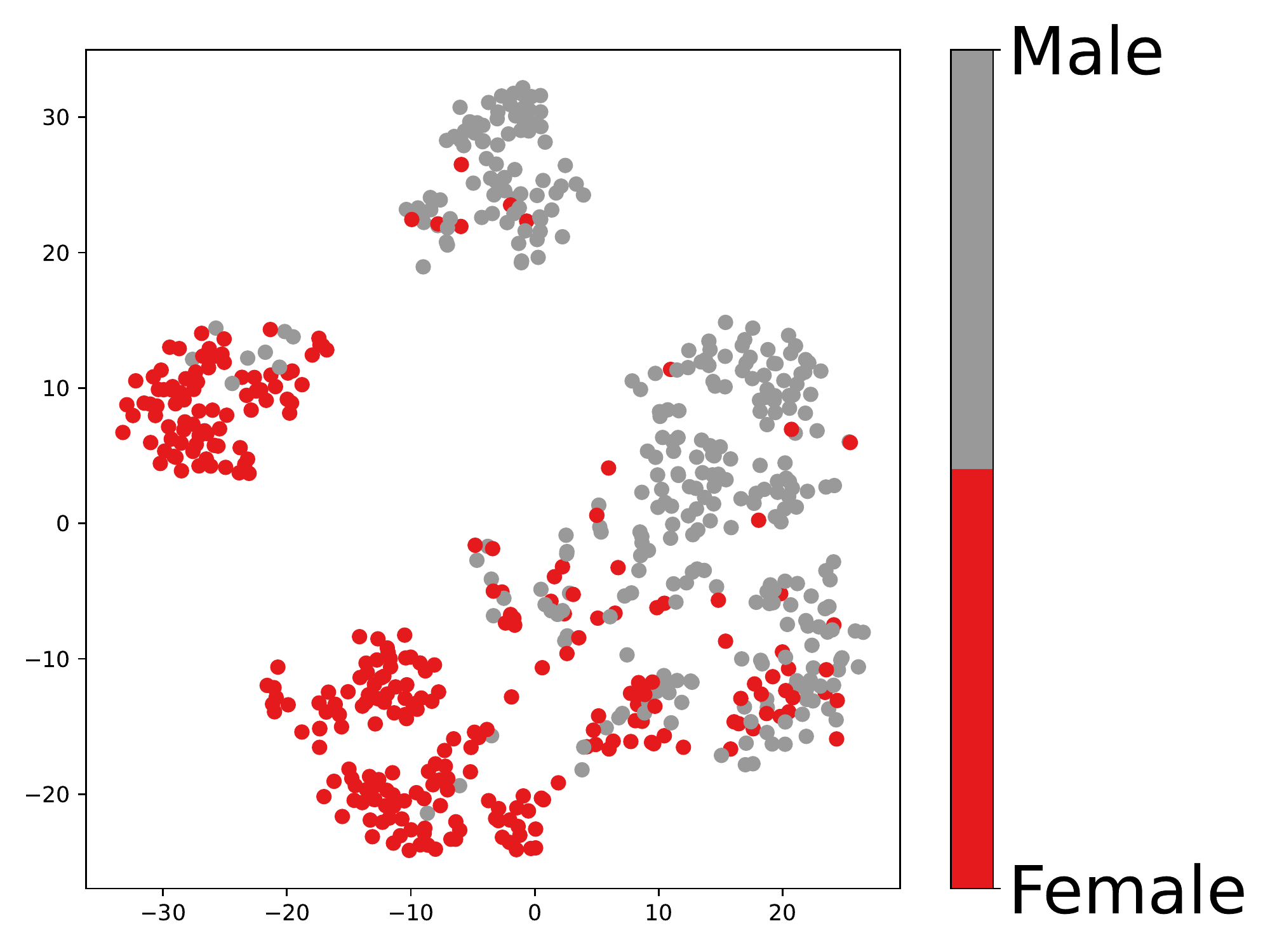}
		\caption{Gender}
		\label{fig:srv_gender_roberta}
	\end{subfigure}
        \hfill
        \begin{subfigure}[]{0.24\linewidth}
		\centering
            \includegraphics[width=\linewidth]{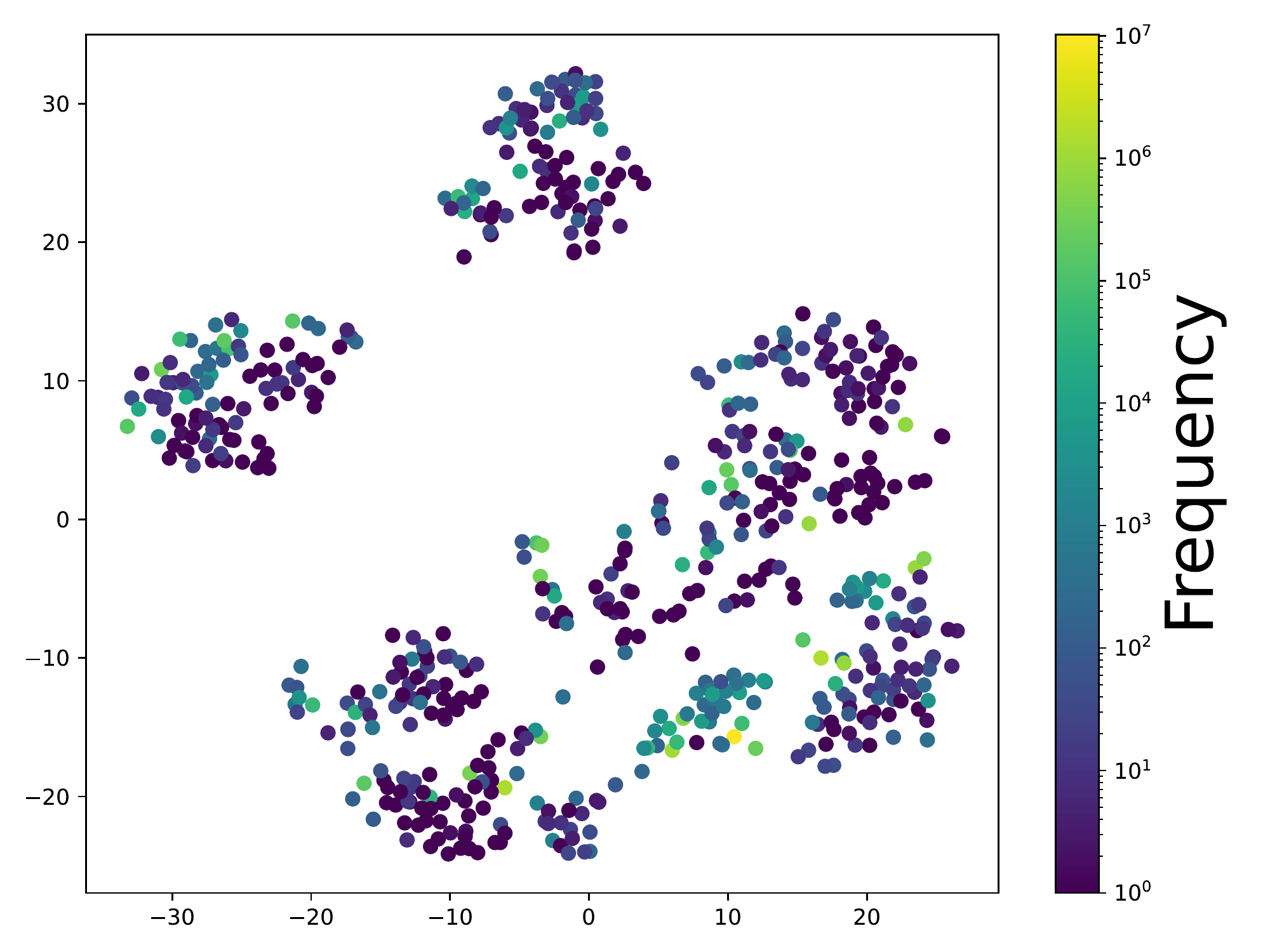}
		\caption{Frequency}
		\label{fig:srv_freq_roberta}
	\end{subfigure}

	\caption{tSNE projections of SR vectors for 608 random names using RoBERTa, visualized by frequency in the pre-training corpus, tokenization length, race/ethnicity, and gender associated with the names respectively. 
        }
	\label{fig:srv_roberta}
\end{figure*}

\begin{figure*}[t]
	\centering
	\begin{subfigure}[]{0.24\linewidth}
		\centering
             \includegraphics[width=\linewidth]{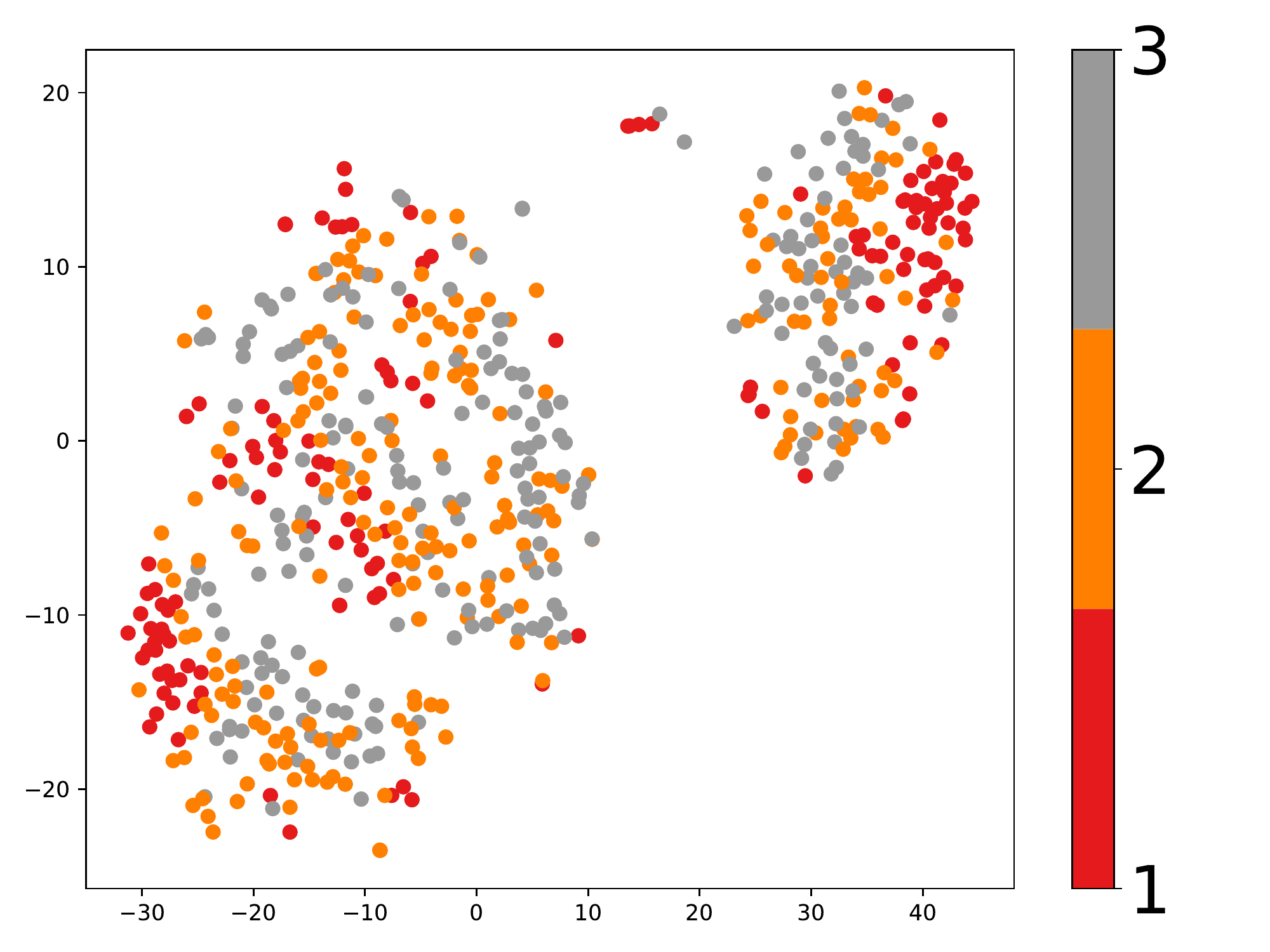}
		\caption{Tokenization length}
		\label{fig:srv_tokenlen_gpt}
	\end{subfigure} 
	\hfill
	\begin{subfigure}[]{0.245\linewidth}
		\centering
            \includegraphics[width=\linewidth]{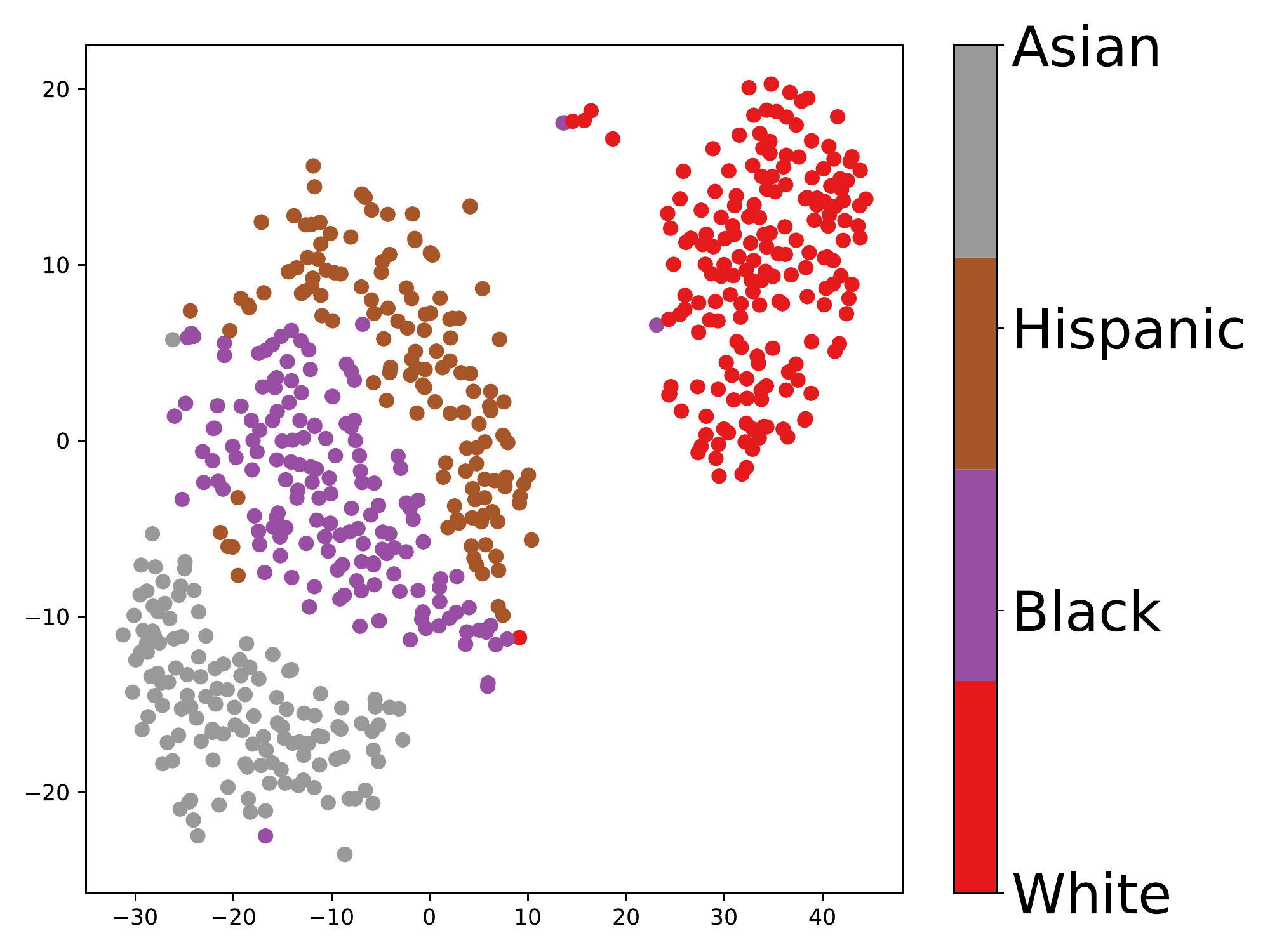}
		\caption{Race/ethnicity}
		\label{fig:srv_race_gpt}
	\end{subfigure}
        \hfill
	\begin{subfigure}[]{0.245\linewidth}
		\centering
            \includegraphics[width=\linewidth]{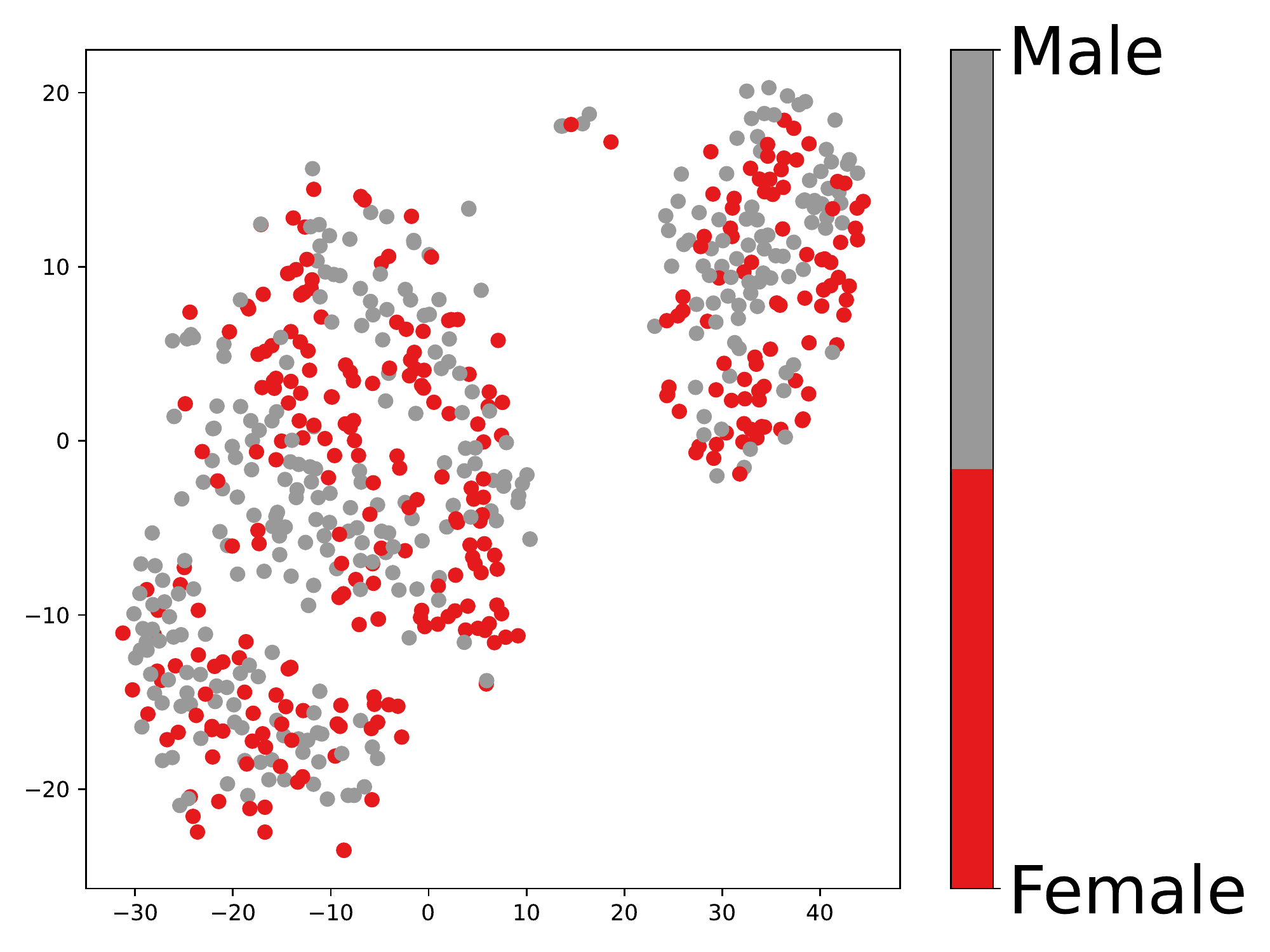}
		\caption{Gender}
		\label{fig:srv_gender_gpt}
	\end{subfigure}
        \hfill
        \begin{subfigure}[]{0.24\linewidth}
		\centering
            \includegraphics[width=\linewidth]{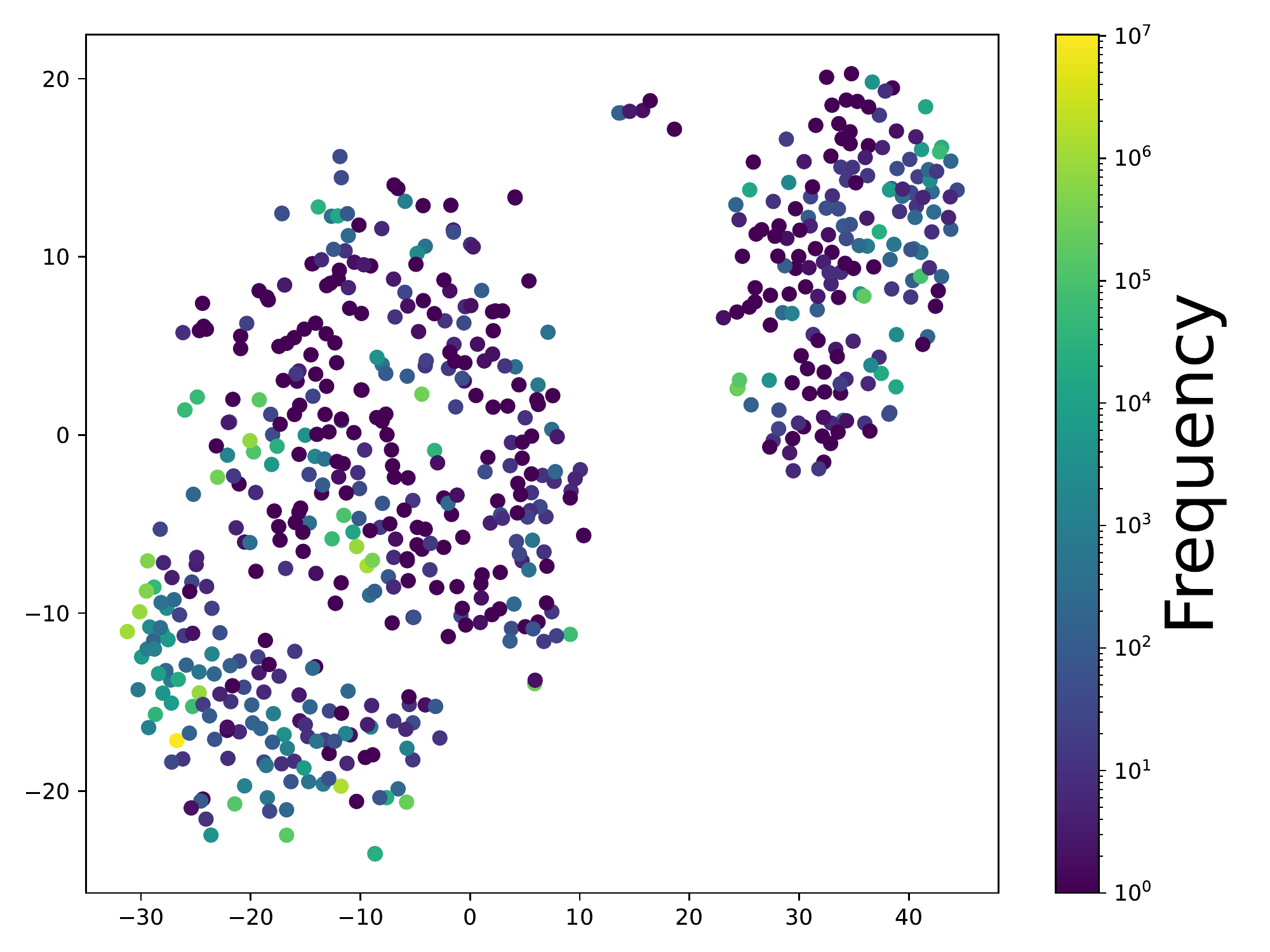}
		\caption{Frequency}
		\label{fig:srv_freq_gpt}
	\end{subfigure}

	\caption{tSNE projections of SR vectors for 608 random names using GPT-2, visualized by frequency in the pre-training corpus, tokenization length, race/ethnicity, and gender associated with the names respectively. 
                }
	\label{fig:srv_gpt}
\end{figure*}

We illustrate the tSNE projections of SR vectors for RoBERTa and GPT-2 in Fig.~\ref{fig:srv_roberta} and Fig.~\ref{fig:srv_gpt} respectively. 
The dimension of the SR vectors is $660$ for these two models.
The plots show that, as we control each of the factors in our analysis, both RoBERTa and GPT-2 treat names differently in the downstream task of social commonsense reasoning. 

\begin{figure*}[t]
	\centering
	\begin{subfigure}[]{0.47\linewidth}
		\centering
		\includegraphics[width=\linewidth]{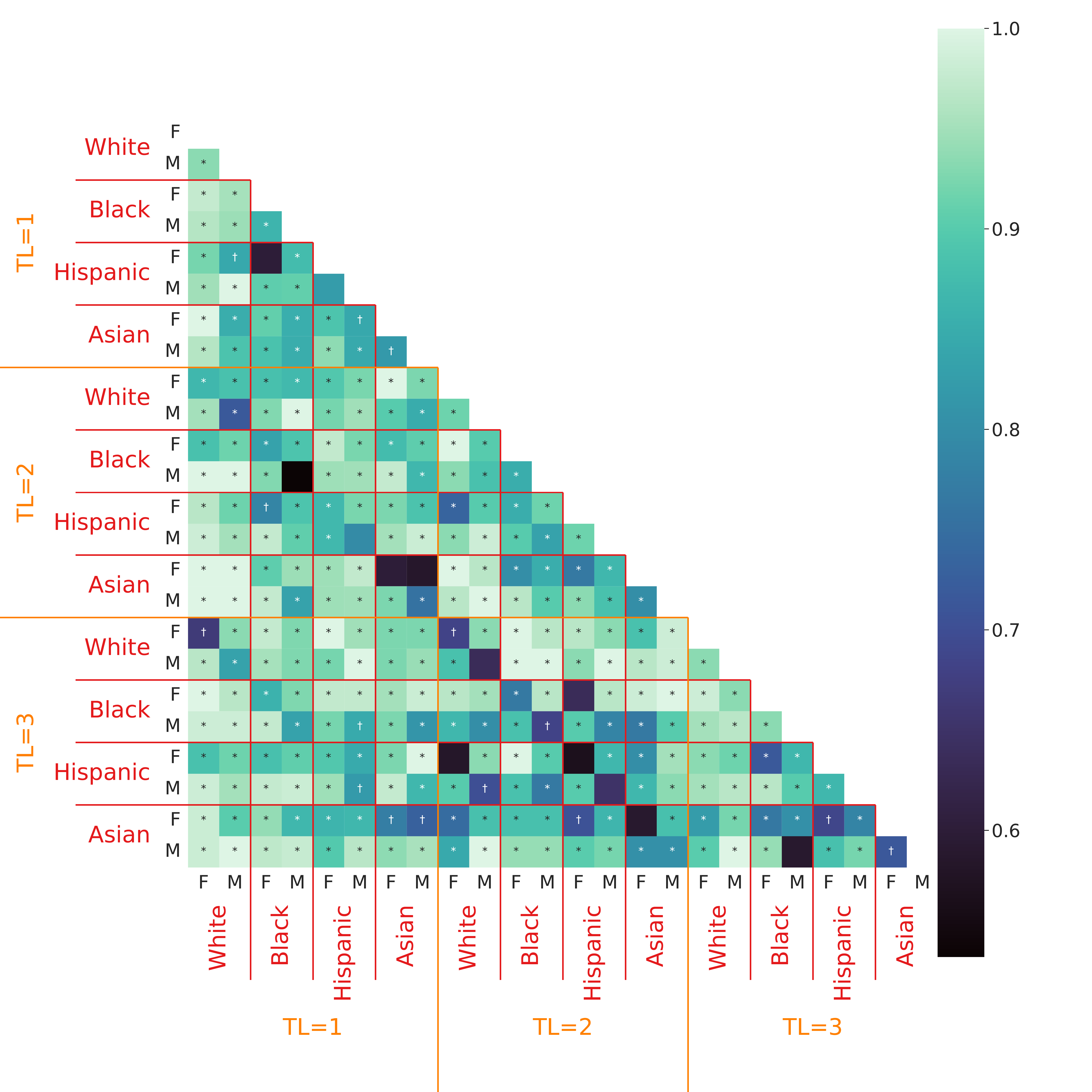}
		\caption{RoBERTa}
		\label{fig:mem_heatmap_roberta}
	\end{subfigure}
	\hfill
	\begin{subfigure}[]{0.47\linewidth}
		\centering
		\includegraphics[width=\linewidth]{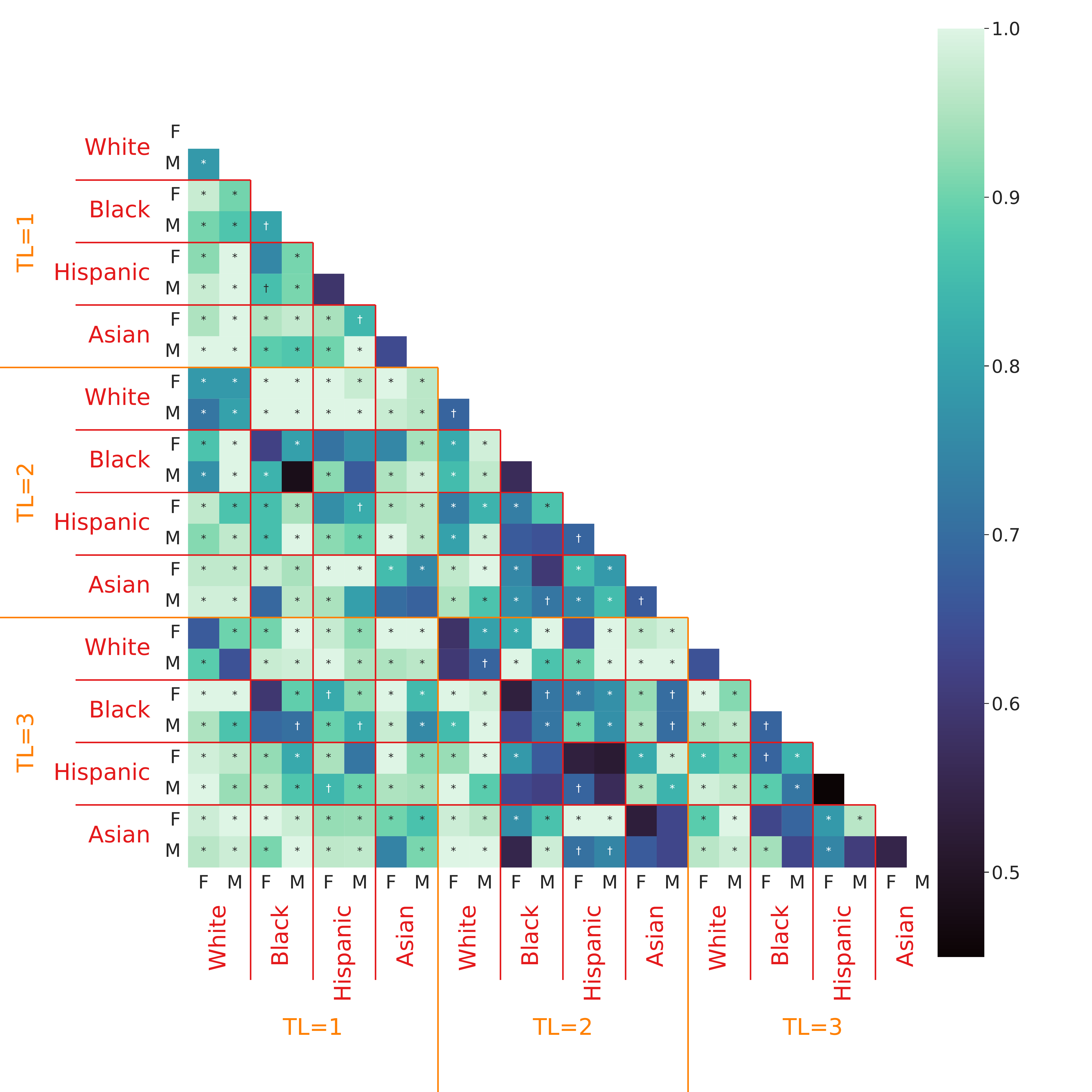}
		\caption{GPT-2}
		\label{fig:mem_heatmap_gpt}
	\end{subfigure}

	\caption{
 Membership prediction accuracy of SR vectors (pairwise comparisons). An ideal accuracy is close to $0.5$. ``TL'': tokenization length. ``F'': female. ``M'': male. * indicates statistical significance at $p < 0.001$ and $\dagger$ at $p < 0.01$.
    }
\label{fig:mem_heatmap_appendix}
\end{figure*}

We also report the membership prediction accuracy for RoBERTa and GPT-2 in Fig.~\ref{fig:mem_heatmap_appendix}.
We observe that gender, race/ethnicity, and tokenization length are all strongly correlated with the model's disparate treatment of names in these models as well.
GPT-2 behaves similarly as BERT, where tokenization length, race/ethnicity, and gender are all factors that indicate the model's different behavior towards names.

\section{Responsible NLP}
\paragraph{Licenses}
We have used BERT, RoBERTa, and GPT-2 for our empirical studies.
BERT uses Apache License Version 2.0,\footnote{\url{https://www.apache.org/licenses/LICENSE-2.0}} and both RoBERTa and GPT-2 use MIT License.\footnote{\url{https://opensource.org/licenses/MIT}} We are granted permission to use and modify these models for our experiments per these licenses.

We also use Stanford NER in our experiments, which is under GNU General Public License (V2 or later).\footnote{\url{https://www.gnu.org/licenses/old-licenses/gpl-2.0.html}}

The pipeline SODAPOP is under Attribution-NonCommercial-NoDerivatives 4.0 International (CC BY-NC-ND 4.0).\footnote{\url{https://creativecommons.org/licenses/by-nc-nd/4.0/}} We have the permission to copy and redistribute the material in any medium or format.

The dataset Social IQa is under Creative Commons Attribution 4.0 International License\footnote{\url{https://creativecommons.org/licenses/by/4.0/}} as it was published by Association for Computational Linguistics. Per the license, we may ``copy and redistribute the material in any medium or format'' and ``remix, transform, and build upon the material for any purpose, even commercially.''

The first name dataset~\cite{rosenman2022race} is under CC0 1.0 Universal (CC0 1.0) Public Domain Dedication.\footnote{\url{https://creativecommons.org/publicdomain/zero/1.0/}} Everyone can copy, modify, distribute and perform the work, even for commercial purposes, all without asking permission. 

\paragraph{Consistency with the intended use of all artifacts}
We declare that the use of all models, datasets, or scientific artifacts in this paper aligns with their intended use.